\documentclass[acmsmall,screen]{acmart}
\setcopyright{none}
\makeatletter
\renewcommand\@formatdoi[1]{\ignorespaces}
\makeatother
\definecolor{colorXuchao}{RGB}{255,0,0}   
\definecolor{colorChetan}{RGB}{0,200,255}   
\definecolor{colorDylan}{RGB}{0,0,255}   

\usepackage{amsmath}
\usepackage{bbm, dsfont}
\usepackage{subcaption}
\usepackage{mdframed}
\newcounter{newfinding}

\usepackage{tcolorbox}
\usepackage{booktabs,colortbl}
\definecolor{lightblue}{rgb}{0.0, 0.2, 1.0}
\colorlet{transblue}{lightblue!20}

\newcommand{\qone}{Can we accurately estimate the confidence for the root cause recommendations generated by Large language models?}
\newcommand{\qtwo}{Given the importance of domain expertise in cloud incident root cause analysis, do historical records improve the calibration of the confidence estimates?}
\newcommand{\qthree}{Can the model leverage its own analyses to generate more accurately calibrated scores?}
\newcommand{\qfour}{Can our proposed approach generalize to encompass different root cause analysis models?}
\newcommand{\toolname}{LM-PACE}
\newcommand{\tooltext}{\textbf{P}rompting and \textbf{A}ugmentation for \textbf{C}onfidence \textbf{E}stimation with \textbf{L}anguage \textbf{M}odels}

\tcbuselibrary{listings,breakable}
\begin{document}

\title{PACE-LM: Prompting and Augmentation for Calibrated Confidence Estimation with GPT-4 in Cloud Incident Root Cause Analysis}
\author{Shizhuo Dylan Zhang}
\email{shizhuo2@illinois.edu}
\affiliation{%
  \institution{University of Illinois Urbana-Champaign}
  \city{Champaign}
  \state{Illinois}
  \country{USA}
}
\author{Xuchao Zhang}
\email{xuchaozhang@microsoft.com}
\affiliation{%
  \institution{Microsoft}
  \city{Redmond}
  \state{Washington}
  \country{USA}
}
\author{Chetan Bansal}
\email{chetanb@microsoft.com}
\affiliation{%
  \institution{Microsoft}
  \city{Redmond}
  \state{Washington}
  \country{USA}
}
\author{Pedro Las-Casas}
\email{pedrobr@microsoft.com}
\affiliation{%
  \institution{Microsoft}
  \city{Redmond}
  \state{Washington}
  \country{USA}
}
\author{Rodrigo Fonseca}
\email{Fonseca.Rodrigo@microsoft.com}
\affiliation{%
  \institution{Microsoft}
  \city{Redmond}
  \state{Washington}
  \country{USA}
}
\author{Saravan Rajmohan}
\email{saravan.rajmohan@microsoft.com}
\affiliation{%
  \institution{Microsoft}
  \city{Redmond}
  \state{Washington}
  \country{USA}
}


\begin{abstract}
Major cloud providers have employed advanced AI-based solutions like large language models to aid humans in identifying the root causes of cloud incidents. Despite the growing prevalence of AI-driven assistants in the root cause analysis process, their effectiveness in assisting on-call engineers is constrained by low accuracy due to the intrinsic difficulty of the task, a propensity for LLM-based approaches to hallucinate, and difficulties in distinguishing these well-disguised hallucinations. To address this challenge, we propose to perform confidence estimation for the predictions to help on-call engineers make decisions on whether to adopt the model prediction. Considering the black-box nature of many LLM-based root cause predictors, fine-tuning or temperature-scaling-based approaches are inapplicable. We therefore design an innovative confidence estimation framework based on prompting retrieval-augmented large language models (LLMs) that demand a minimal amount of information from the root cause predictor. This approach consists of two scoring phases: the LLM-based confidence estimator first evaluates its confidence in making judgments in the face of the current incident that reflects its ``grounded-ness" level in reference data, then rates the root cause prediction based on historical references. An optimization step combines these two scores for a final confidence assignment. We show that our method is able to produce calibrated confidence estimates for predicted root causes, validate the usefulness of retrieved historical data and the prompting strategy as well as the generalizability across different root cause prediction models. Our study takes an important move towards reliably and effectively embedding LLMs into cloud incident management systems.

\end{abstract}

\maketitle

\section{Introduction}
In the past decade, the IT industry has been progressively adopting cloud platforms as their preferred method for deploying applications and services. Within these expansive cloud services, incidents like unexpected interruptions or performance declines can substantially harm customer satisfaction, leading to revenue loss and a decrease in customer confidence. Root Cause Analysis (RCA), as a critical task in the incident diagnosis process, aims to uncover the underlying causes of issues and implement corrective measures to prevent future occurrences. Identifying the incident's root cause is crucial for reducing service downtime, mitigating customer impact, and minimizing manual effort.



\begin{figure}
    \centering
    \includegraphics[width=\textwidth]{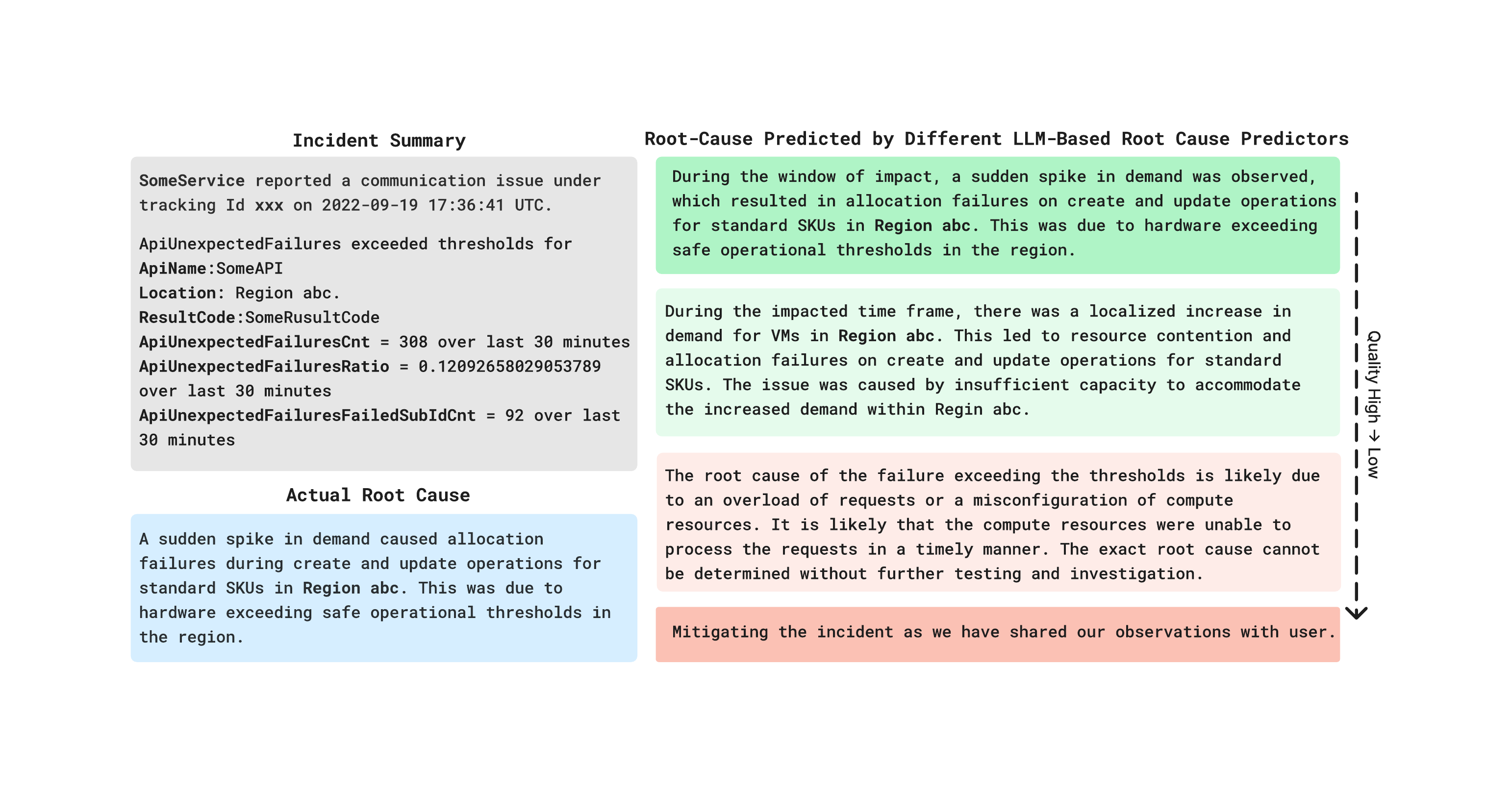}
    \vspace{-15mm}
    \caption{An example cloud incident with a series of predicted root causes by different root cause predictor models. This illustrative example shows that the model predictions are of mixed quality even for the same incident. The strength of green (red) color for model-predicted root causes indicates how likely they are correct (wrong). It is apparent that if such information can be obtained beforehand, we could significantly assist on-call engineers in making decisions on how much should they read through and rely on each recommendation.}
    \label{fig:examples}
\end{figure}

Recently, Large Language Models (LLMs) have demonstrated their proficiency in incident diagnosis tasks, including root cause analysis. They play a crucial role in gathering information from historical data and various sources, providing invaluable recommendations to on-call engineers (OCEs) when addressing intricate cloud incidents. While LLMs can deliver promising outcomes in numerous situations, they still generate a significant number of incorrect recommendations. These inaccuracies have the potential to misguide OCEs into making erroneous decisions, leading to a waste of their diagnostic efforts. As shown in the example depicted in Figure~\ref{fig:examples}, there are situations where LLMs can provide invaluable insights, reducing the need for exhaustive report analysis, delving into past incidents, or data gathering. However, in other instances, they may appear vague, lacking in detail, or even misleading. Moreover, LLM-based approaches are susceptible to generating hallucinations. What's concerning is that recognizing these hallucinations can be a challenging task, making it difficult for engineers to determine their authenticity. These issues significantly impede the incorporation of AI tools into the cloud incident management process, as erroneous predictions can lead engineers astray, resulting in significant time wastage and potentially serious consequences for both the service provider and its users.


In order to address the gap in the practical application of AI-assisted Root Cause Analysis (RCA), we propose the solution to estimate the confidence level associated with each predicted root cause. By establishing a reliable confidence estimation mechanism, on-call engineers can make more informed decisions. This, in turn, enhances the efficiency and effectiveness of incident resolution while reducing the costs incurred by delays caused by inaccurate root cause recommendations from the model. Moreover, preventing the adoption of incorrect recommendations not only improves the user experience by minimizing disruptions but also reduces the on-call effort required to rectify such issues. By gaining a deeper understanding of the reliability of AI-based model output, organizations can confidently integrate these tools into their operational workflows, ensuring seamless and reliable service delivery in cloud environments.

However, developing a calibration procedure for root-cause analysis is itself challenging, encompassing a range of factors that complicate the process. First, the root cause predictors might vary across time and services. Our solution needs to be general enough and decoupled from the underlying model. Second, the access to the model can be limited. For example, the logits and probabilities of GPT-3.5-Turbo and GPT-4 models are not disclosed, making it difficult to apply fine-tuning-based approaches or traditional post-hoc~\cite{jiang2021knowitknow} methods like temperature scaling to directly obtain confidence scores from the root cause predictor. Further, the open-ended nature of the problem adds to the challenge. Compared with categorical tasks like multiple-choice or classification, calibration for open-ended question-answering tasks is more complex in nature due to the unconstrained output space, and difficulty in defining direct supervision signals. 
This paper aims to make progress in resolving the critical challenge of confidence calibration in domain-specific applications, with a focus on root cause analysis of cloud incidents. The calibration problem involves evaluating the plausibility and reliability of the outputs generated by AI-empowered cloud incident root cause analysis systems, which requires both comprehension of the incident and predicted root cause, and the critical analysis of the prediction. To this end, we adopt a large language model (LLM)-based solution to perform confidence estimation.

Nevertheless, it is worth emphasizing that LLMs are primarily trained on a general-domain corpus from the internet, while the handling of cloud incidents and root causes requires domain knowledge. There is no surface-level connection between an incident description to its root cause, and the model therefore is not able to deduce to the conclusion on the extent to which a root cause is valid by utilizing only its general-domain knowledge and commonsense logical reasoning capacities. To bridge this gap and enhance the performance of the confidence estimator in this domain-specific application, we seek to enrich the LLM's domain expertise in this application by incorporating historical incidents and corresponding root causes as reference data for the model to perform more reliable and grounded confidence estimation.

To this end, \textit{\textbf{we present a new framework for confidence estimation}}. Our method uses retrieval-augmented large language models (LLMs) to generate calibrated confidence scores for model-generated root causes and are introduced as \toolname\space(\tooltext). Our proposed method unfolds in two stages. First, to obtain a notion of the confidence estimator's uncertainty in examining the predicted root cause, we prompt the confidence estimator LLM to gauge the strength of evidence drawn from historical references to assess the current incidents. Then, we prompt the model to evaluate the candidate root cause generated from the root-cause predictor based on the references. Finally, we perform an optimization step to find the most suitable confidence assignment combining the two scores.
Through experimental evaluation, we demonstrate that the model is able to verbalize its confidence based on the evidence and its own analysis that can be effectively converted into a well-calibrated score. 
We answer the following research questions in the subsequent sections:\\
\textbf{RQ1:} \qone \\
\textbf{RQ2:} \qtwo \\
\textbf{RQ3:} \qthree \\
\textbf{RQ4:} \qfour

Through this research, we seek to narrow the gap in the adoption of LLM-based cloud incident root cause predictors, which aid human engineers by providing not just predictions but also well-calibrated confidence estimates. Our method makes minimal assumptions on underlying root cause predictors. We envision our general framework being adaptable to various AI-assisted IT operations scenarios, as long as they share similar task configurations and data availability.


\section{Background}
\subsection{Prospects and Challenges of Large Language Models}
\label{sec:llm}
Pretrained extensively on text corpora from the internet, large language models (LLMs) have demonstrated impressive capabilities in various tasks within the NLP domain~\cite{bubeck2023sparks} and beyond~\cite{deng2023large,zhang2023getting}. Recent advances in instruction tuning and reinforcement learning from human feedback of those models are shown to effectively improve their capabilities in following complex instructions~\cite{Ziegler2019humanpref,bang2023multitask,ouyang2022instructions}. 

To adapt LLMs to a specific down-stream task, one can either fine-tune (i.e. modify the weights of the pre-trained model by performing gradient updates) on the data of this task or perform in-context learning by showing examples or verbally explaining the task without updating the model parameters.

Despite being good at language understanding and text generation, the key for LLMs to successfully answer queries and help humans in various tasks is knowledge~\cite{izacard2022retrievallm}. A variety of works have attempted to improve the performance of LLMs on different tasks~\cite{mialon2023augmentedsurvey} by equipping it with external knowledge. One class of approaches that is particularly interesting under the context of domain-specific tasks is retrieval augmentation, where the relevant text chunks are retrieved by a retriever model based on relevance to the query and presented to the model in the input~\cite{wang2023industrydomainqa,guu2020retrieval,shi2023replug}. The LLMs, though lacking that knowledge during the training phase, can comprehend the retrieved information for its use in generating responses.

While LLMs have exhibited advanced capabilities to produce seemingly coherent and contextually appropriate responses, they have been observed to frequently "hallucinate" \cite{rawte2023surveyhallucinate,zhang2023sirenshallucinatesurvey,ye2023hallucinate,dhuliawala2023hallucinatechainofverification}, producing content that deviates from the original input, contradicts earlier context, or invents information. These inconsistencies, well disguised by their eloquence outputs language models, are subtle and may not be readily discernible to humans. This behavior is particularly pronounced when the model lacks knowledge upon which to ground its answers. One prominent example of such a situation would be domain-specific applications like AIOps.


\subsection{AI-Assisted Root Cause Analysis and Troubleshooting for Cloud Incidents}



\begin{figure*}
    \includegraphics[width=0.8\textwidth]{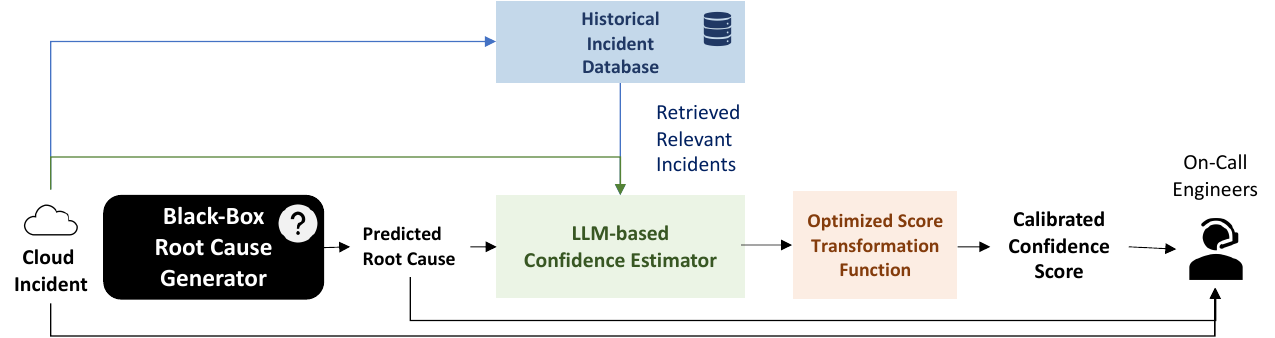}
    \caption{High-level overview of retrieval-augmented confidence calibration.}
    \label{fig:simple_overview}
\end{figure*}

Cloud incident management is a significant challenge for cloud service providers, as incidents can result from various issues like code bugs, dependency failures, and configuration errors. Identifying the root cause is crucial to resolving the incident. Traditionally, on-call engineers have been responsible for this task. They analyze incidents by extracting key information from descriptions and referencing past events and reports.

The rise of LLMs offers an opportunity to loop AI-based systems in root cause identification. Accurate AI predictions can save engineers considerable time, eliminating the need to manually review extensive documentation. However, if AI predictions are incorrect, it can lead to wasted efforts and potential service disruptions. Engineers face a dilemma: blindly trusting AI can be risky, but ignoring all the model predictions means missing out on its benefits.
\begin{table}[!htp]\centering
\small
\begin{tabular}{lcc}\toprule
\textbf{RCA Model} &\textbf{RCA-Strategy} &\textbf{\% Correctness } \\\midrule
\textbf{GPT-4} &Prompting + 8000 Tokens of Most Relevant Historical Incidents &43.40\% \\
\rowcolor{transblue}
\textbf{GPT-3.5-Turbo} &Prompting + 3896 Tokens of Most Relevant Historical Incidents &22.30\% \\
\textbf{GPT-3.5-Turbo} &Prompting + 3 Most Relevant Historical Incidents &23.70\% \\
\rowcolor{transblue}
\textbf{Text DaVinci-003} &Prompting + 3896 Tokens of Most Relevant Historical Incidents&30.50\% \\
\textbf{Text DaVinci-003} &Prompting + 3 Most Relevant Historical Incidents &25.60\% \\
\bottomrule
\end{tabular}
\caption{Correctness rates of the root cause predictors on cloud incidents.}\label{tab: correctness}
\end{table}


As shown in Table~\ref{tab: correctness}, though a reasonable percentage of the model-generated root causes are correct given the difficulty of the task (43.4\% for GPT-4; around or less than 30\% for GPT-3.x series), in which cases the troubleshooting time can be significantly shortened, whereas in the rest of the cases believing the erroneous model-produced root causes can cause delays in resolving the issue and restoring services. Another layer of complication comes from the indiscernible wrong predictions.  Figure \ref{fig:misleading} shows an example of an incident and the root causes generated by GPT-4 model compared with its actual root cause as diagnosed by on-call engineers. They are pointing to very different issues, and following the hint from the model-predicted root cause would only lead the engineers farther away from mitigating this issue. However, due to the complex nature of cloud services, engineers would not be able to tell immediately whether it is correct at first sight based on their own knowledge - since the model predictions are usually correct-looking. Therefore they need assistance when deciding whether to follow the model-generated root cause.
Hence, one of the most crucial steps in developing an AI-assisted root cause analysis system lies in effectively quantifying the level of trustworthiness associated with each model-generated root cause for an incident. It ensures the seamless integration of AI technologies in cloud incident management and fosters improved decision-making and incident resolution. 

\subsection{Confidence Calibration}
Confidence calibration refers to the problem of predicting probability estimates that represent the
true correctness likelihood. Formally, for any pair of input $X_i$ and the model-predicted output $\hat{y}_i$ indexed by $i$ in the dataset, the confidence estimate for the model prediction $\tilde{P}_i (\hat{y}_i|X_i)$ should satisfy
\begin{equation}
    \mathbbm{P}(l_i| \tilde{P}_i = p) = p , \forall p \in [0,1]
\end{equation}, where we use $l_i$ to denote the event that $i$-th data point is predicted correctly. Verbally, the meaning of the equation would be `` the probability of a data point with confidence $\tilde{P}_i = p$ to be correct should be $p$". Note that the definition we adopted here is generalized to incorporate different types of prediction tasks, including open-ended question-answering, compared to the classic definition for classification where $\hat{y}_i, y_i \in Y$ are label classes in the label space, and
$l_i = \mathbbm{1} ( \hat{y}_i = y_i)$. In practice, we often adopt a binning-based strategy where we divide the probability interval of 0 to 1 into equal buckets, To estimate the calibrated probability ($p$), we calculate the proportion of actual positive outcomes that correspond to the predicted probabilities within each bucket.

To quantify the result of calibration, a commonly used metric is Estimated Calibration Error (ECE), defined as
\begin{equation}
\text{ECE} = \sum_{m=1}^M \frac{|B_m|}{n} \left| \text{acc}(B_m) - \text{conf}(B_m) \right|
\end{equation} where \(M\) is the total number of bins, \(B_m\) is the set of indices of samples that fall into bin \(m\), \(n\) is the total number of samples, \(\text{acc}(B_m)\) is the accuracy of bin \(m\) calculated as the proportion of correct predictions in bin \(m\), and \(\text{conf}(B_m)\) is the confidence of bin \(m\) calculated as the average predicted probability in bin \(m\).

The issue of confidence estimation attains heightened significance in the context of LLMs due to their direct engagement with human users in multiple scenarios. As elaborated in Section~\ref{sec:llm}, as LLMs are used to perform tasks of higher complexity, the imperative for assessing the grounding of these models’ predictions in concrete knowledge, as well as the soundness of their reasoning processes, also increases. This necessitates the development of strategies to effectively measure the confidence of the model outputs.

Model interpretability gains paramount significance in the symbiosis of human and machine learning systems. Cosmides and Tooby \cite{cosmides1996cognitive} highlighted the innate human ability to comprehend probabilities, underscoring the value of transparent, interpretable confidence estimates. Such clarity allows humans to establish an informed level of trust in model outputs, thus fostering enhanced collaboration with machine learning systems. By integrating estimated confidence levels with predictions, individuals can adeptly assess the reliability of the model's responses, paving the way for informed decision-making and effective human-machine interaction.

However, confidence calibration in open-ended question-answering is inherently more challenging than in classification, multiple-choice, or extractive question-answering. In these other formats, a limited, predefined set of responses makes confidence calibration easier. However, the infinite possibilities in open-ended tasks make this process more complex. An additional layer of difficulty appears when there is potential utilization of external information since the model is only aggregating information without possessing it, rendering it incapable of intrinsically evaluating the accuracy of its responses without careful design.

\section{Problem Definition}
In this work, we propose to overcome the difficulties in adopting LLM-based tools to assist humans by confidence estimation. We devise methods to produce calibrated confidence scores for model-predicted root causes for cloud incidents to assist on-call engineers in deciding the extent to trust the model-recommended root cause. 

Confidence estimation for recommended root causes presents two critical challenges that demand careful consideration. The first aspect revolves around the necessity for confidence estimation methods that are both highly adaptable and broadly applicable. Given that the settings for root cause predictors - like the model type used for predictions, the volume of accessible information, and others - are diverse across services and evolving over time, the confidence estimation technique should be effective irrespective of the root cause predictor in place. This adaptability ensures that the confidence estimation remains reliable across various models and services. Consequently, it is essential for the confidence estimator to operate independently of the root cause predictor, ensuring its adaptability and accuracy under varying circumstances.

Moreover, although language models like GPT-3.5-Turbo and GPT-4 exhibit high efficiency in diverse applications, essential data, including model parameters (weights), probabilities, and logits for output tokens, remain unavailable. These elements are fundamental for confidence calibration. The unavailability of weights obstructs the possibility of fine-tuning the model to align confidence, while logits and probability scores are indispensable for implementing post-hoc calibration techniques. This limitation complicates the development of dependable confidence estimation procedures and demands an approach that does not require the above information.

Therefore, a realistic assumption is to treat the root cause generator as a black box where we only assume access to its generated text, $\hat{r}_{query}$ without knowing anything else, including the configurations, weights, and output probabilities/logits of the model, the algorithm used to produce the output, prompt, or any auxiliary information used by the model. Also, the choice of root cause generator can encompass any model that produces open-ended textual responses as its predicted root causes. This unrestricted perspective accounts for the diversity and flexibility of root cause generation models, accommodating a wide spectrum of approaches. Given the incident description $d_{query}$ and the predicted root cause ${\hat{r}}_{query}$, our goal is to return a confidence estimate $\psi$ that reflects the confidence of ${\hat{r}}_{query}$ being the correct root cause. 

To this end, our goal is to develop a stand-alone confidence estimator that remains decoupled from the root cause analysis procedure. By doing so, we can ensure the versatility and adaptability of our confidence estimation approach, enabling it to work seamlessly across different language models while overcoming the limitations of restricted accessibility to certain language models. In pursuit of this, we need to answer several research questions: \\
\textbf{RQ1: \qone }\\
As previously established, in situations where the root cause generator operates as a black box, there arises a necessity for an independent mechanism, incorporating a language model. This module serves the purpose of comprehending the root causes predicted by the generator and generating outputs that can be subsequently converted into confidence scores. A fundamental question is the feasibility of achieving well-calibrated confidence estimation with a prompting-based approach as well as how to appropriately design such a framework.


\textbf{RQ2: \qtwo}\\
Root cause analysis, together with many other industry application scenarios of LLMs, poses the same challenge of lacking domain-specific knowledge. In particular, when requiring the model to analyze the root cause of an incident critically, it necessarily needs to draw from relevant knowledge - which, due to the domain-specific nature, is not available from training. We draw inspiration from the various prior works on in-context retrieval-augmentation for large language models and see if the model can benefit from relevant historical incidents to produce more accurate confidence estimates.

\textbf{RQ3:\qthree}\\
Performing a rating/scoring task, especially in such an open-ended situation, is non-standard and non-trivial for a large language model. Therefore, we explore whether providing the model with guidance to undertake a preliminary qualitative evaluation in textual form before proceeding to assign scores can prove advantageous in this particular context. This approach aims to investigate the potential efficacy of enhancing the model's performance by incorporating an intermediate step of the textual assessment, which could potentially address the complexities associated with accurate scoring in such an unconstrained setting.

\textbf{RQ4:\qfour}\\
As previously highlighted, a pivotal consideration revolves around ensuring the method's generalizability to different root-cause distributions from a variety of root-cause analysis models. To address this concern, we systematically evaluate the efficacy of the proposed approach across different sets of root causes generated by various models. By subjecting our method to this comprehensive testing, we aim to ascertain its robustness and versatility in accommodating varying characteristics of root-cause predictions.

\section{Methodology}
\begin{figure*}
    \centering
    \includegraphics[width=.9\textwidth]{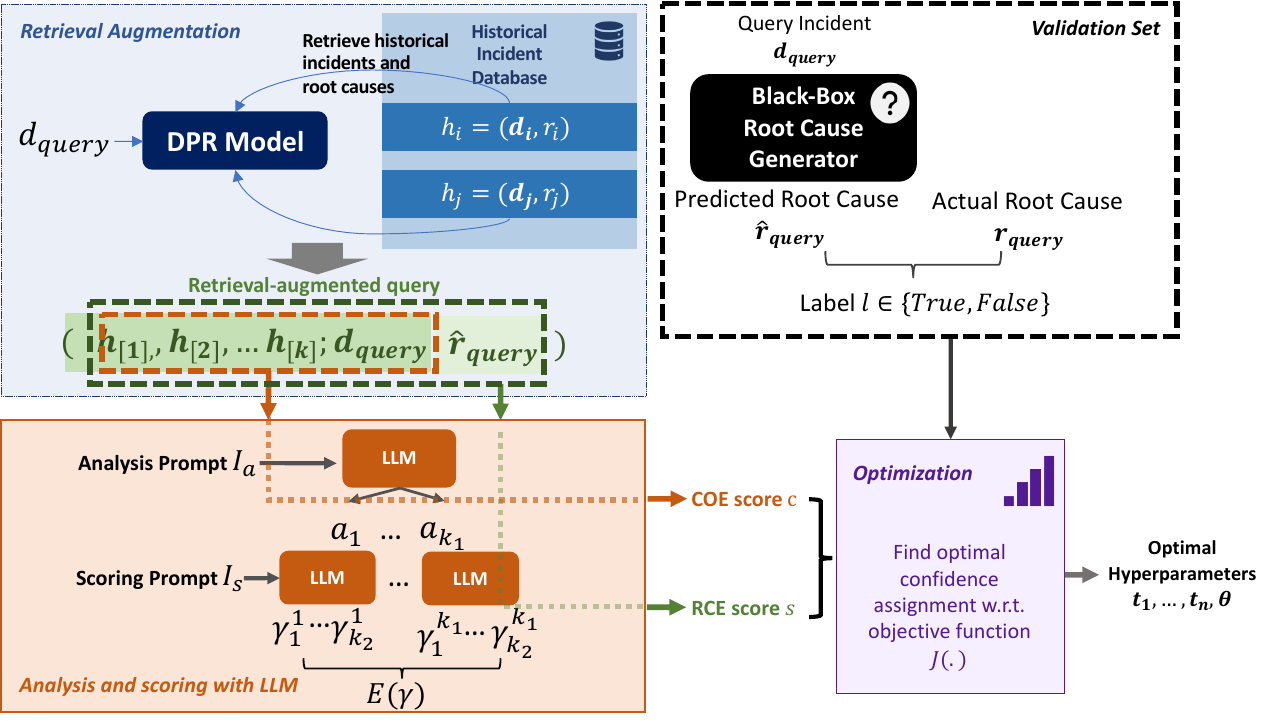}
    \caption{An overview of the with \toolname\space calibration method. Note that we are using different analysis prompts and scoring prompts for COE-scoring and RCE-scoring respectively.}
    \label{fig:overview}
\end{figure*}

In our research, we introduce a retrieval-augmented two-step procedure designed to conduct confidence estimation for model-generated cloud incident root causes, leveraging the capabilities of large language models through prompting. The proposed approach involves multiple stages to ensure reliable and accurate results.

To prepare the retrieval-augmented data, we retrieve a list of similar historical incidents relevant to each query. This retrieval process is done by employing semantic similarity-based dense retrievers, which enable us to identify past incidents with potential connections to the current problem at hand.

Following the retrieval of historical incidents, we proceed to prompt the large language model to perform confidence-of-evaluation (COE) pre-examination. During this stage, the model is given the historical incidents, and their associated root causes, along with the description of the current incident. It is then prompted to consider whether it possesses sufficient information to analyze the underlying cause of the current incident.

\begin{figure}[htb]
  \centering.
\begin{tcolorbox}[title=Example Incident With Model Generating Misleading Root Cause]
\textbf{Incident Description}\\
All machines in \texttt{SomeService} that tries to upgrade OS fails and become unreachable ... we have test VM's created where the issue reproduces consistently. The VM comes up healthy. Then after reboot VM goes into the problematic state.\\ 
\textbf{Model-Generated Root Cause}\\
The machines in \texttt{SomeService} faced OS upgrade failures due to incompatibility with the current machine firmware. When the VMs rebooted post-upgrade, the outdated firmware caused a conflict, rendering them unreachable. Upgrading the machine firmware to the latest version is essential to ensure compatibility and successful OS upgrades.\\
\textbf{Actual Root Cause} \\
The function \texttt{KiAdjustGroupConfiguration} can cause an uncommitted subnode to be placed in a committed group. The current \texttt{KiAllocateProcessorNumber} does not account for this, leading it to recommit the subnode's group, which reassigns new group numbers to all subnodes in that group. This disrupts the group database, causing synchronization issues and potential errors during boot, especially during the process of ideal node assignment.

\end{tcolorbox}
  \caption{An example of a misleading root cause prediction. It is difficult to decide whether the model-generated root cause is pointing to the correct troubleshooting direction. }
  \label{fig:misleading}
\end{figure}
Next, we perform root cause evaluation (RCE) utilizing the information from the historical incidents. In this step, we ask the language model to evaluate the generated root cause of the current incident based on the retrieved historical incidents and their root causes. Finally, to obtain the final confidence score for each root cause, we combine the COE and RCE scores together and perform an optimization step. 
By adopting this retrieval-augmented procedure, we aim to establish a robust framework for confidence estimation in cloud incident root causes. This innovative approach empowers us to leverage the capabilities of large language models effectively, thereby enhancing our ability to make informed decisions and troubleshoot cloud-related incidents with increased accuracy and confidence.


\subsection{Retrieval-Augmented Confidence Calibration}
One of the fundamental obstacles in confidence estimation for the root cause analysis task stems from its strongly domain-specific characteristics. Off-the-shelf large language models are primarily designed for general-domain applications and lack the necessary domain-specific knowledge required for a specific service. On the other hand, relevant historical incidents and the expert-recommended root causes can offer crucial insights into the current incidents' root cause. 

Therefore, we implemented a retrieval-based pipeline, utilizing a retriever module to search for pertinent historical incidents to enhance the confidence estimator model.

Concretely, we adopted dense retrieval. Given a database $D = \{h_i = (d_i,r_i):i=1,2,....n{_{max}}\}$  containing historical incidents $h_i$ each of which is a pair of historical incident description $d_i$ and its ground truth root cause $r_i$,
for each query incident $d_{query}$, we compute the inner product $\langle {Enc}(d_{query};\theta_{enc}),{Enc}(d_{i};\theta_{enc}) \rangle$ for each $d_i \in D$, where we first compute the embeddings of the descriptions of query instances and instances in the retrieval database $D$ using the same encoder LM $Enc$ parametrized by $\theta_{enc}$ and then obtain their similarity score by calculating the inner product of their encoded vectors. 

Then, for each $d_{query}$, we retrieve a sorted list of most relevant historical incidents from the database based on the similarity scores up to a fixed token budget $L$ to form the retrieved reference $H$ for the query, that is, we retrieve the $k$ most relevant examples where
$k = max({k'})$ $s.t.$ $\mathbf{len}([h_{[1]},...,h_{[k']}]) \le L $
where $h_{[j]}$ is the $j$-th highest ranked incident with respect to $d_{query}$ and $\mathbf{len}(.)$ returns the total number of tokens in a list of data instances. Note that for clarity of notations, we omitted the superscripts indexing the query incident in this subsection. 

\begin{figure*}
    \centering
    \includegraphics[width=.8\textwidth]{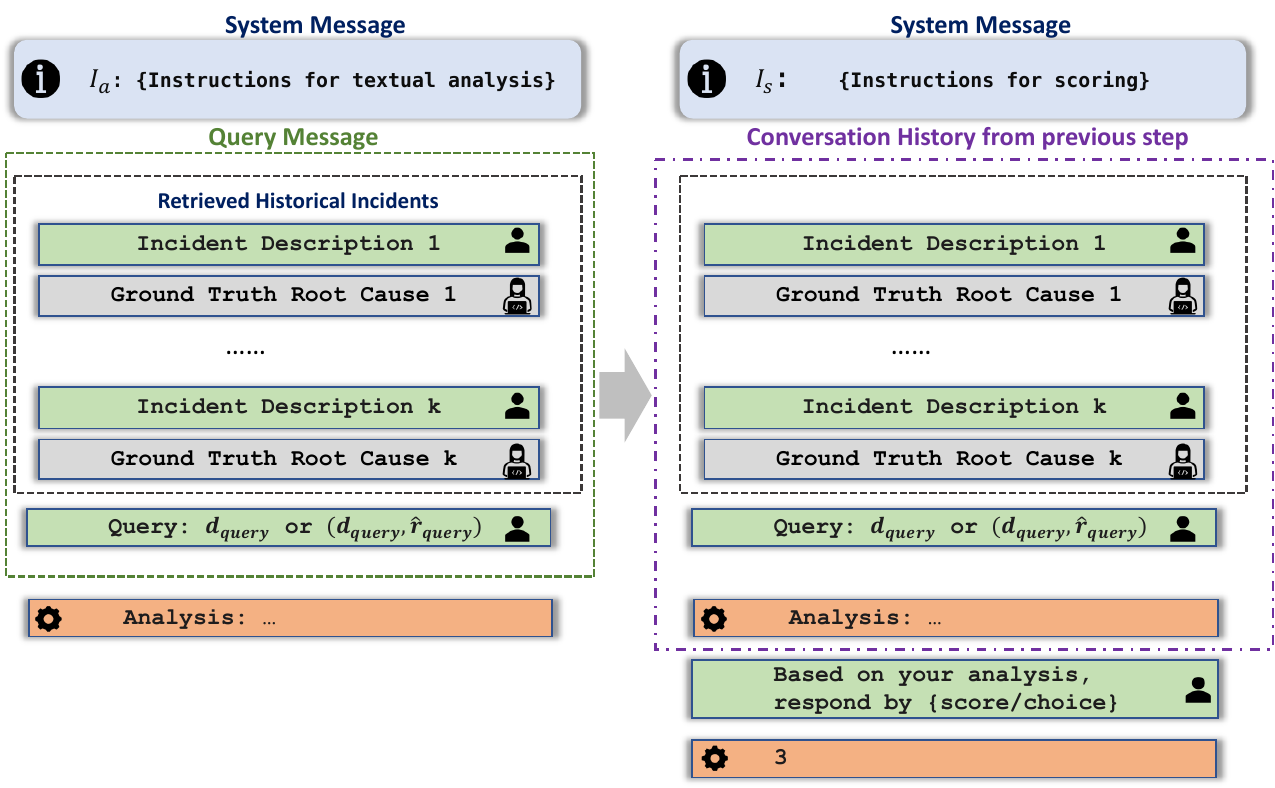}
    \caption{Illustration of prompting-based scoring procedure applied in both COE and RCE steps.}
    \label{fig:prompting}
\end{figure*}
\subsection{Confidence-of-Evaluation (COE) Pre-examination}
The model's assessment of an incident's root cause relies significantly on historical incidents. However, the historical incidents retrieved using the semantic similarity-based retriever might encompass a range of situations, differing in relevance to the current incident. Consequently, these incidents might not offer sufficient guidance to the model for precisely evaluating the root cause of the given query incident. Additionally, given the recognized occurrence of hallucination in large language models as discussed in Section~\ref{sec:llm}, there is a varying degree of reliability in the model's judgments. Therefore, this step aims to address the varying levels of hallucination and assess the extent to which the model's evaluation of the root cause is based on factual and reliable information.

The confidence-of-evaluation (COE) estimator takes in the retrieved historical incidents $H = \{h_1,...h_k\} $ ( where each $h_i = (d_i,r_i)$ is a pair of historical incident descriptions and its ground truth root cause) together with $d_{query}$ which is the query incident, and decides whether we have enough evidence from $H$ to reason about the root cause of the current incident. The goal of this step is to obtain the model's level of confidence in its ability to reason effectively about the root cause of the query incident given the retrieved information. If the model is low in confidence of determining the root cause due to the lack of information, its evaluation of the generated root cause would also be less trustworthy.


As observed in works studying problem-solving with LLMs~\cite{nye2021scratchpad,dhuliawala2023hallucinatechainofverification} the LLMs benefit from simulating the process of articulating their intermediate cognitive processes while addressing a task. Through the verbalization of thought processes, the model gains the ability to meticulously plan its approach to a question, elucidate intricate steps of reasoning, deconstruct complex problems into manageable sub-problems, and retain and utilize partial deduction outcomes and interim computation results. Ultimately, this method empowers the model to arrive at a well-considered and logical final decision.

In our scenario, given that both deciding on the retrieved incidents' suggestive power and quantifying the confidence of model-predicted root causes are inherently complicated and multifaceted in nature, we seek to elicit the model's ability to make judgments by inducing it to produce intermediate analysis.

Concretely, as demonstrated in Figure~\ref{fig:prompting}, the model first needs to generate analysis in textual form conditioning on the historical incidents, description of the current incident, and the instruction for analysis $I{_{a}^{COE}}$. In order to reduce biases and cover more facets, we sample multiple analyses for each query~\cite{li2023making}. $a{_{ij}^c} \sim p(a | H, d_{query}, I{_{a}^{COE}})$ where $j=1,2,...,k_1$ and $k_1$ is the number of analyses for each incident.

Then we sample $k_2$ binary responses from model on whether the model thinks the historical conditioning on each analysis, that is, for each $a{_i^c}$, we sample responses $c{_j^i} \sim p(c |H, d_{query}, I{_{s}^{COE}}, a{_i^c})$ and estimate the confidence-of-evaluation (COE) score with the empirical mean across all scores obtained from all analyses $$E(c) = \frac{1}{k_1 \times k_2}\sum_{j=1,...,{k_2}}\sum_{k=1,...,{k_1}} c{_j^i}$$.  In this step, we provide the model with a multiple-choice question consisting of two options (\textbf{\textit{Yes/No}}) and ask it to pick the letter of choice and $c{_j^i}= \mathbbm{1}(\text{choice}==\textbf{yes}) \in \{0,1\}$.


\subsection{Root Cause Evaluation (RCE) Scoring}

In the second step of the confidence estimation, we ask the model to evaluate the candidate's root cause based on the historical incidents. 
The root cause evaluator first takes in the same $H = \{h_1,...h_k\}$ again as in COE step, together with $(d_{query},{\hat{r}}_{query})$ where ${\hat{r}}_{query}$ is the predicted root cause to be evaluated, to analyze whether the predicted root cause ${\hat{r}}_{query}$ is a plausible root cause to the query incident based on historical records. Again, we first ask the model to provide with analysis $a{_{ij}^s}$'s following certain rubrics we specified in the instruction $I{_{a}^{RCE}}$, and then ask the model to score following the instruction $I{_{s}^{RCE}}$. 

Similar to the COE step, we sample ${k_1}'$ analyses for each incident.
Yet, we take on a different scoring mechanism. The evaluation of root cause is more complicated as there are multiple dimensions including truthfulness (i.e. whether it contains false information), groundedness (i.e. to what extent the historical incidents support or go against the generated root cause), and informativeness (i.e. the level of detail in the generated root cause, and its adequacy in directing engineers for troubleshooting, relative to the guidance offered for analogous historical incidents). Therefore, instead of simply asking for binary responses, we prompt the model to produce scores on a specified scale conditioning on each analysis. 
The score is generated from  $s{_j^i} \sim p(s |H, d_{query},{\hat{r}}_{query}, I{_{s}^{RCE}}, a{_i^s})$. Similar to step-1, we again aggregate the mean score  $$E(s) = \frac{1}{{k_1}' \times {k_2}'}\sum_{j=1,...,{k_2}'}\sum_{k=1,...,{k_1}'} s{_j^i}$$ where we again ask the model to produce ${k_2}'$ scores conditioning on each analysis.


\subsection{Estimating Confidence From COE and RCE Scores}
Given the COE and RCE scores obtained from two-step confidence estimation for each data point, the problem remains to be finding an optimal mapping from these intermediate scores into confidence.

Consider a problem where we assumed $m$ categories of confidence level evenly dividing the interval $[0,1]$, and that
want to assign each root cause the most appropriate category that best indicates the confidence level. Let $\pi(.,.)$ denote the score transformation function that takes in COE and RCE scores and maps them into a value between 0 and 1, and $\omega(.)$ be the weighting function that determines the relative importance of each category, $\mathbbm{1}[l_j]$ denote the event that the generated root cause $j$ is correct, and $t_0,\ldots,t_{m}$ be thresholds for each category where $t_0 = 0 $ and $t_m = 1$ are minimum and maximum possible scores from $\pi(.,.)$, respectively. 

In general, optimization objective takes the form 
\begin{equation}
    \underset{t_0<t_1<\ldots<t_m,\theta}{\text{minimize}} \sum_{i=1}^{m-1} \omega(i) \left|\pi({c_j,s_j}) - \frac{\sum_{j}\mathbbm{1}[t_{i}\le \pi({c_j,s_j})\le t_{i+1}] * \mathbbm{1}[l_j]}{\sum_{j}\mathbbm{1}[t_{i}\le 
 \pi({c_j,s_j})\le t_{i+1}]}\right|
\end{equation}
The parameter $\omega$ plays a pivotal role in adjusting the relative significance attributed to each confidence bucket. For example, in situations where emphasizing high-confidence cases is the top priority, we have the flexibility to allocate greater weightage to the respective bins.
In case of $\omega(i) = \sum_{j}\mathbbm{1}[t_{i}\le \pi({c_j,s_j})\le t_{i+1}]$, the objective reduces to Expected Calibration Error (ECE) score. In our experiment, we take ECE objective, while the weighting mechanism of each bin can be tailored for different scenarios. 

\section{Experiments}

\subsection{Experiment Set-up}
\subsubsection{Dataset and Labels}

In this subsection, we present our dataset preparation process. 

Our collection of data consists of incident data gathered from different services and levels of severity within \textbf{\textit{CompanyX}}, amounting to a total of 121,308 incidents. In order to create this dataset, we chose 98,308 incidents as the retrieval set, 2,000 for the validation set, and the remaining 3,000 for the testing set. We search for the best set of parameters that optimizes the objective (ECE) on the validation set and subsequently perform evaluation on the test set.


For each validation and test instance, we set the maximum token budget $L$ for retrieved incidents to 3896 and retrieve the most similar historical incidents within the budget. The average number of historical incidents is around 15. 

Due to the cost of human labeling, we generated pseudo-labels with GPT-4 by feeding it with pairs of $(r_i,\hat{r_i})$ and ask it whether these are similar. We ask the model to provide with ratings in integer values from 1 to 3. We sample 128 scores per query and query 4 times per input. We compute the average score across the 512 outputs. We decided the threshold for correctness to be $\ge2.3$ which gives the maximum $F1$ score on an independent validation set with human annotations. 
\subsubsection{Evaluation Protocol}

We demonstrate the effectiveness of the method via reliability diagrams, ECE scores, and human evaluation.
Reliability diagrams (\ref{fig:reliability_gpt4} and \ref{fig:reliability_other}) serve as vital tools for evaluating the calibration of probabilistic predictions in machine learning models. A reliability diagram visually compares the predicted probabilities produced by a model against their observed frequencies, offering insights into the model's ability to provide well-calibrated confidence levels. It serves as a rapid assessment of the performance of calibration. In the context of the reliability diagram, the horizontal axis is the confidence axis, representing a range between 0 and 1. This interval is equally divided into smaller sub-intervals. The height of bars in each sub-interval indicates the proportion of positive labels among the collection of root causes that possess an assigned confidence score lying within each respective sub-interval. The regions shaded in gray are intervals within which the corresponding bars are expected to align. 

In addition, we also presented the Expected Calibration Error (ECE) score which stands as a pivotal metric in quantifying the calibration quality of probabilistic predictions generated by machine learning models. ECE gauges the disparity between predicted probabilities and observed frequencies across various confidence intervals. This assessment offers a nuanced perspective on the model's calibration precision, highlighting potential deviations from ideal alignment. By numerically quantifying these discrepancies, ECE facilitates an objective evaluation of a model's reliability in terms of its probability estimates, aiding practitioners in pinpointing and addressing areas requiring calibration enhancement.

\subsubsection{Root Cause Generators}

To assess the efficacy of the proposed approach across diverse black-box models, we conducted experiments involving three distinct language models as root-cause generators that include models fine-tuned both with (GPT-4, GPT-3.5-Turbo) and without human feedback (Text-DaVinci-003) which are expected to produce different output distributions in terms of style, tone, and correctness. These experiments were conducted under various configurations as discussed below.

We explored models exposed to varying degrees of historical information. The annotation \textit{"-3-shot"} denotes root causes generated based on the three most similar historical instances based on semantic similarity of incident descriptions, while the annotation \textit{"-3896"} corresponds to root causes derived from the maximum feasible number of the most similar historical incidents while adhering to a constraint of 3896 tokens for the prompt size. We established a maximum limit of 200 tokens for the root cause prediction. \textit{"-8K"} denotes 8000 tokens of most similar historical incidents.

\subsubsection{Implementation Details}
For conducting confidence estimation in our experiment, we used OpenAI GPT-4 (8K maximum tokens) as our base language model. This powerful model serves as the foundation for assessing the confidence levels associated with each model-generated root cause.
During the process of generating analyses and scoring, we set the temperature parameter ($T$) to 1. The temperature value allows us to control the level of randomness in the model's output. When generating analyses during confidence estimation, we limit the maximum completion token to 256.




\subsection{Experimental Results}
\subsubsection{RQ1: \qone}

Figures \ref{fig:res_full} \ref{fig:davinci3shotfull} \ref{fig:chat3shotfull} \ref{fig:chat4kfull} and \ref{fig:davinci4kfull} are reliability diagrams plotted for the confidence estimates by our full method for root causes generated by various models under different conditions. As demonstrated in the reliability diagrams, our proposed approach is able to produce well-calibrated confidence estimates for the root causes generated under different root cause generator set-ups.

\begin{figure*}
    \centering
    \begin{subfigure}{0.32\textwidth}
        \centering
        \includegraphics[width=\linewidth]{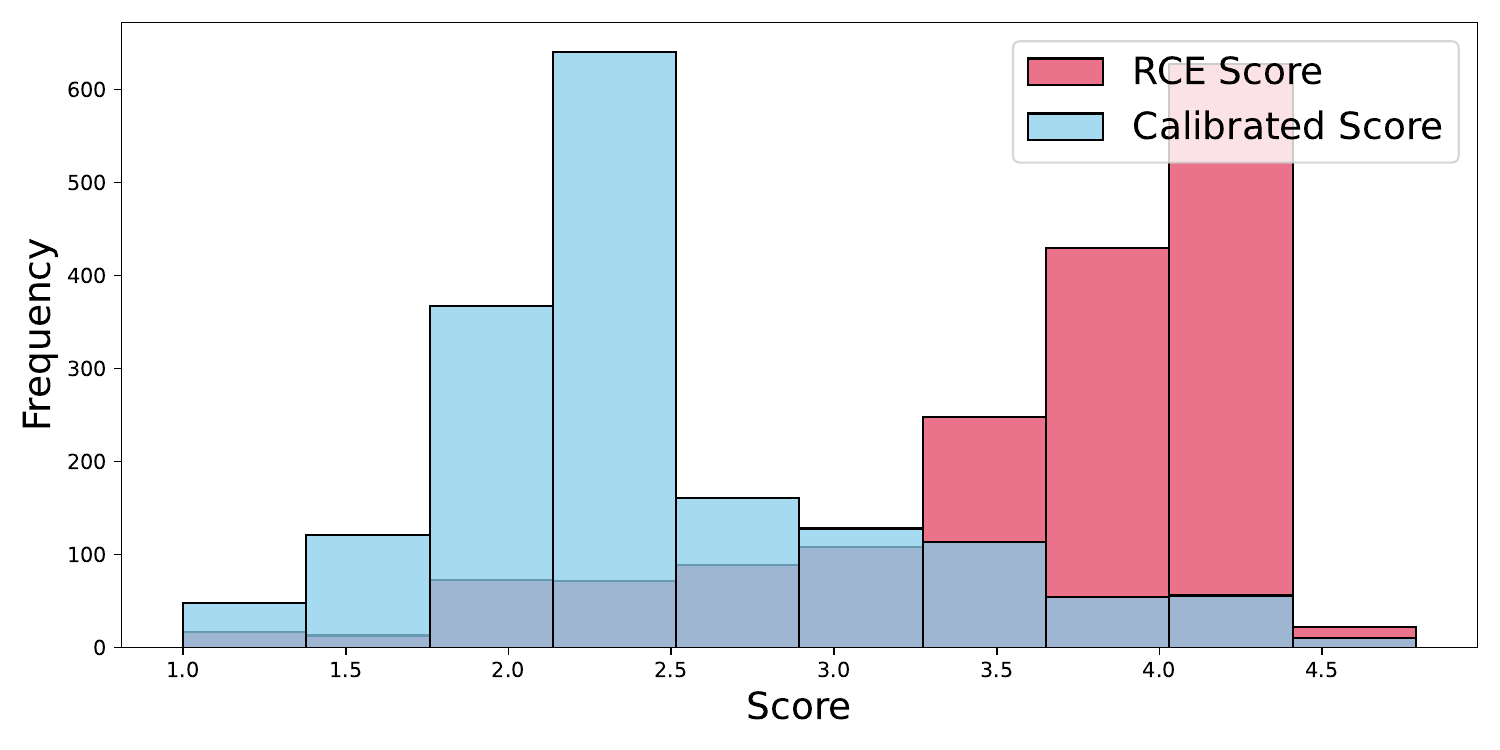}
        \caption{GPT-4-8K.}
        \label{fig:gpt4-score-distribution}
    \end{subfigure}
    \hfill
    \begin{subfigure}{0.32\textwidth}
        \centering
        \includegraphics[width=\linewidth]{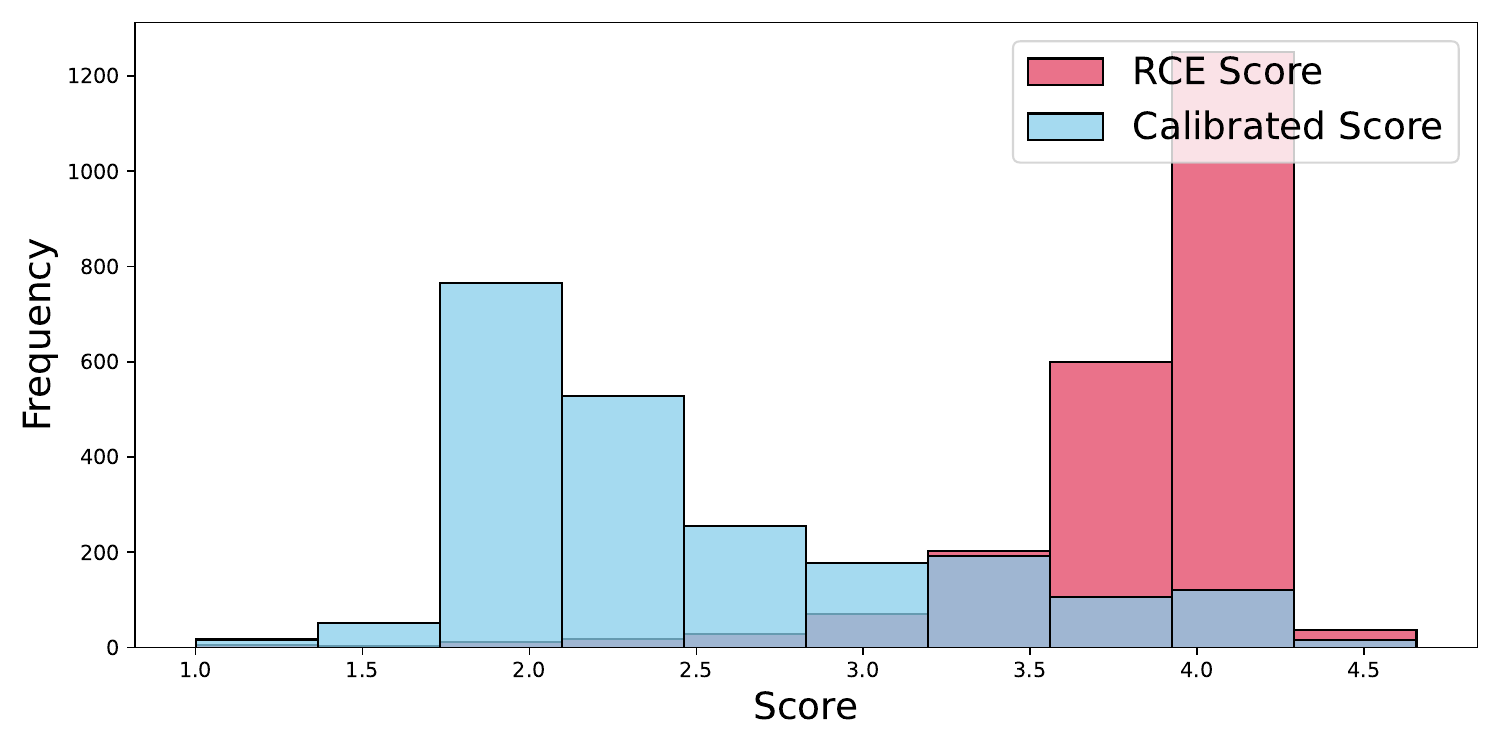}
        \caption{Text-DaVinci-003-3shot.}
        \label{fig:davinci-score-distribution}
    \end{subfigure}
    \hfill
    \begin{subfigure}{0.32\textwidth}
        \centering
        \includegraphics[width=\linewidth]{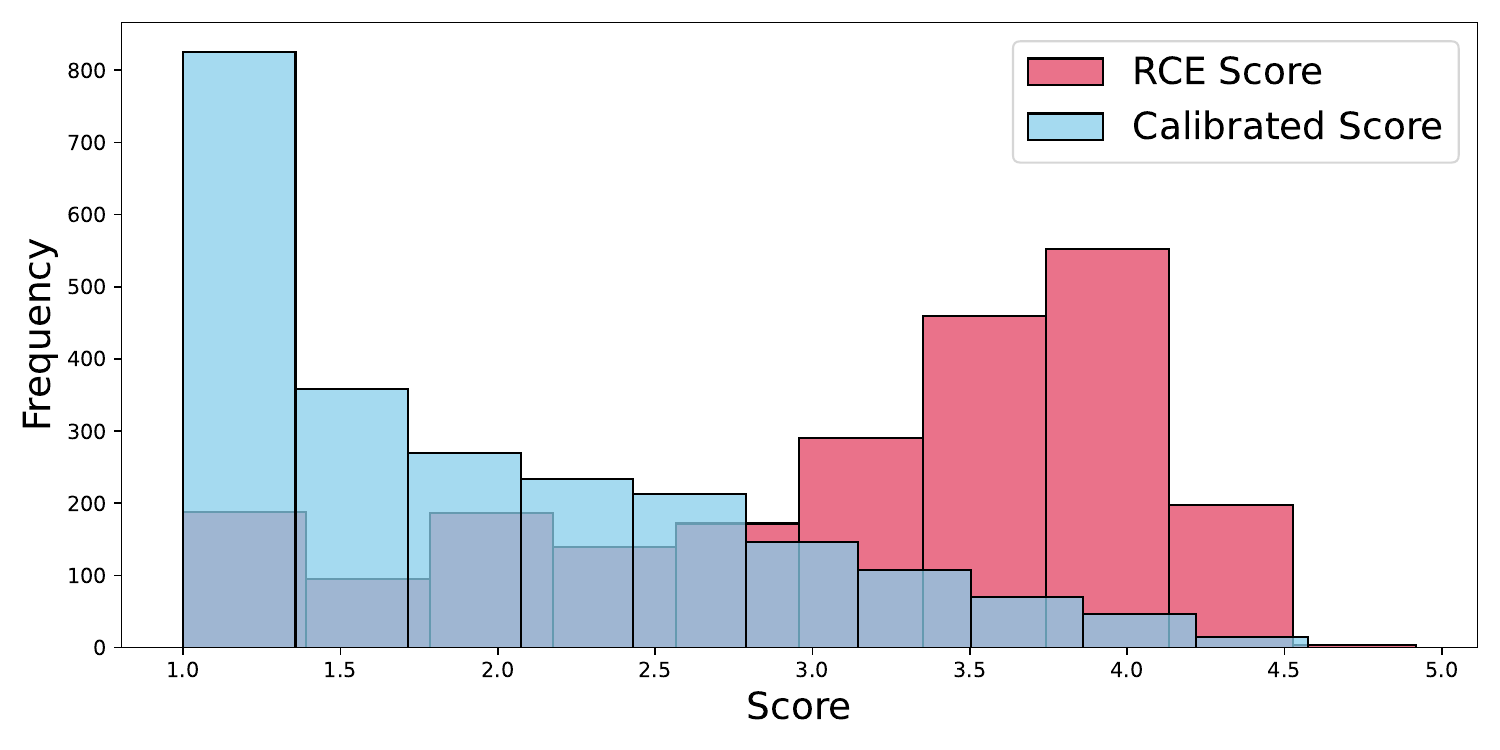}
        \caption{GPT-3.5-Turbo-3shot.}
        \label{fig:turbo-score-distribution}
    \end{subfigure}
    \caption{Frequency histogram of scores for incorrect root cause predictions. Distributions of RCE scores and full confidence score are very different. }
    \label{fig:dis_compare}
\end{figure*}

To assess the significance of the COE score, we conducted the same optimization procedure solely on the basis of the RCE score (see Figures ~\ref{fig:davinci3shot_step2} ~\ref{fig:chat3shot_step2} ~\ref{fig:chat4k_step2}~\ref{fig:davinci4k_step2}). Notably, the confidence scores are seriously mis-calibrated when using only RCE scores. In the confidence estimation procedure, the COE score plays an important role in producing a calibrated confidence score since it counters the score hallucination effect in situations where the calibrator model is not well-informed enough to present a fair evaluation of the root cause. In Figure~\ref{fig:dis_compare}, we juxtapose the distributions of RCE scores with calibrated scores for incorrect root causes. The evident disparity between these two score distributions underscores that relying solely on the RCE score is inadequate for generating well-calibrated confidence estimates. 

We analyze the distributions of cases in which the model assigns high COE scores, indicating instances where it deems itself equipped with ample evidence to engage in reasoning about the underlying factors. This is contrasted with cases receiving low scores. In situations of low COE scores, the model demonstrates a tendency to overestimate incorrect root causes (see Figure \ref{fig:dis_low}), while undervaluing correct root causes (See Figure\ref{fig:dis_high}). This divergence is less pronounced in instances with high COE scores. It implies that the model is indeed more confused when there is less evidence from historical incidents. This observation underscores the vital importance of considering the model's confidence in estimation and integrating such insights into the confidence estimation framework.

\begin{figure*}
    \centering
    \begin{subfigure}[t]{0.24\textwidth}
        \centering
        \includegraphics[width=\linewidth]{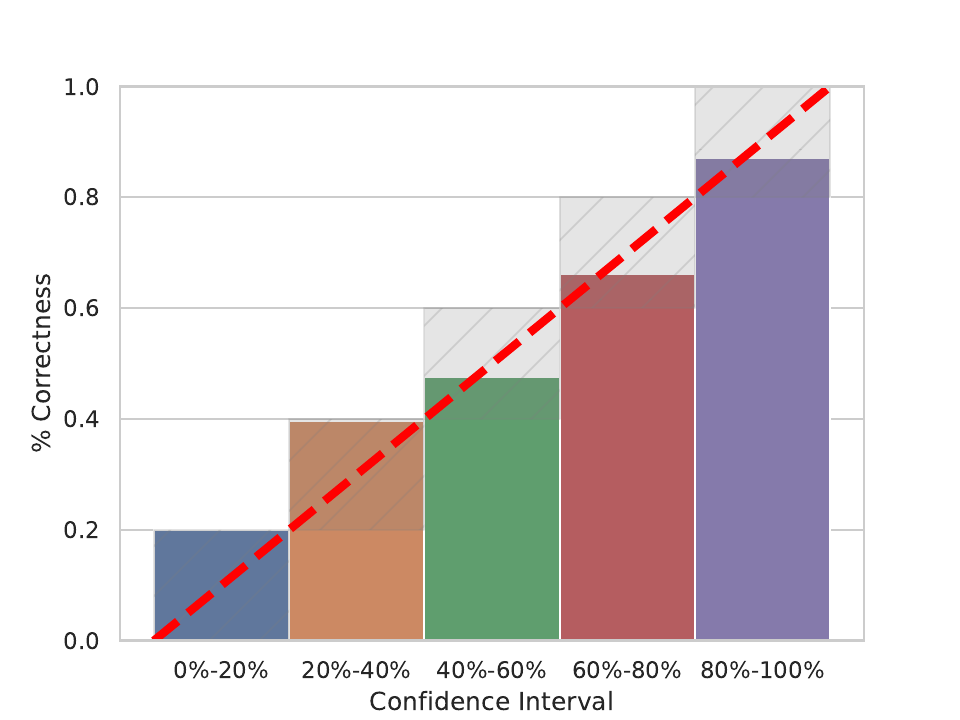}
        \caption{Confidence Estimation by full \toolname}
        \label{fig:res_full}
    \end{subfigure}
    \hfill
    \centering
    \begin{subfigure}[t]{0.24\textwidth}
        \centering
        \includegraphics[width=\linewidth]{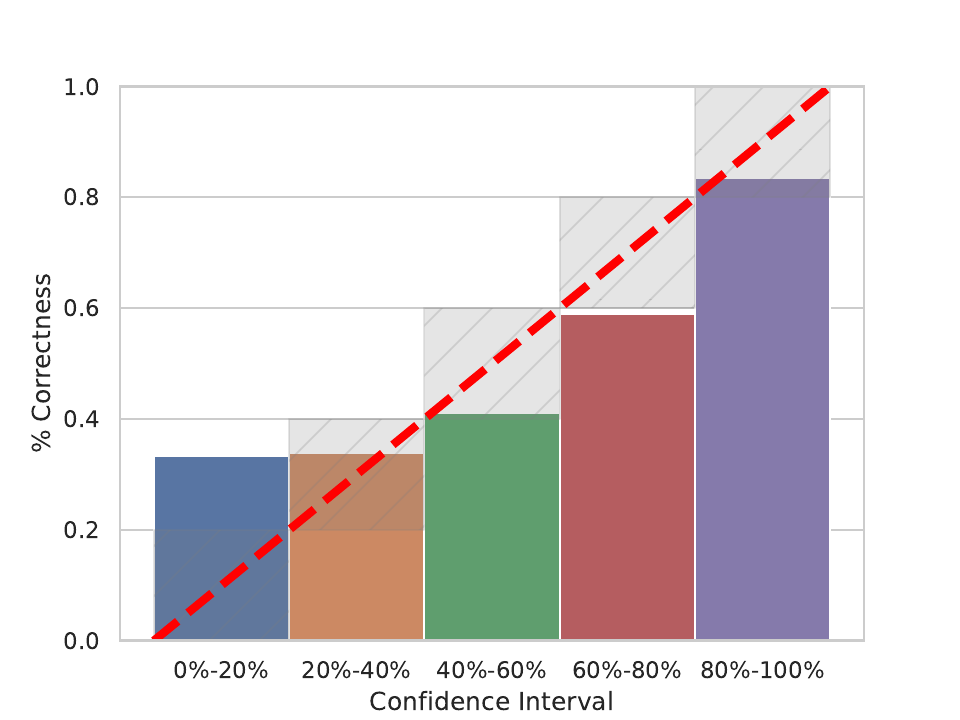}
        \caption{Confidence Estimation Without COE estimation.}
        \label{fig:res_full}
    \end{subfigure}
    \hspace{1mm}
    \begin{subfigure}[t]{0.24\textwidth}
        \centering
        \includegraphics[width=\linewidth]{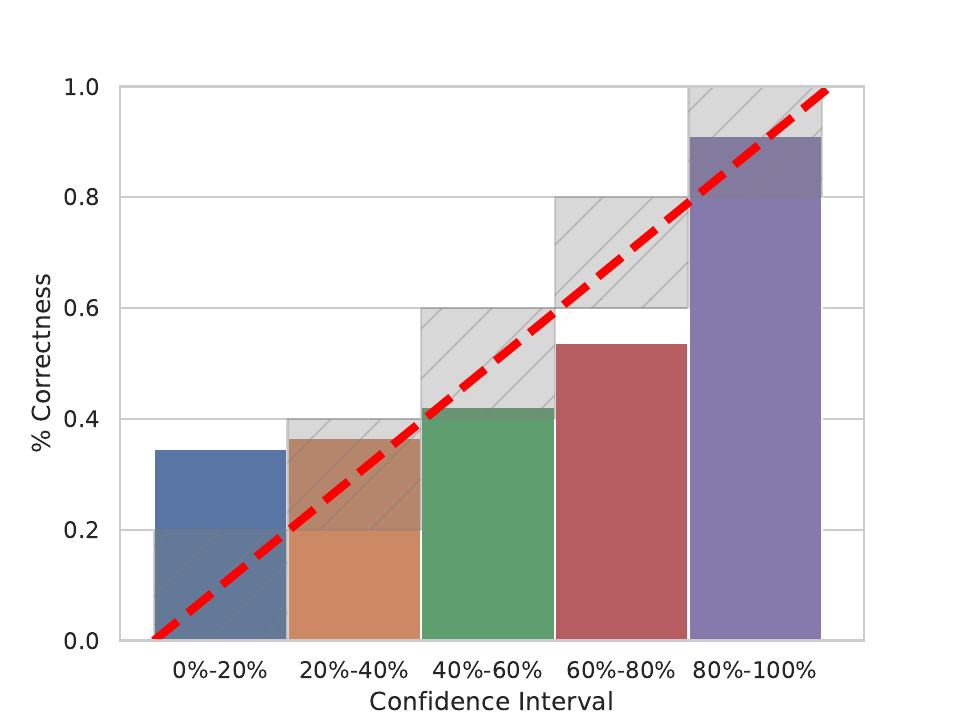}
        \caption{Confidence Estimation by Uniform Binning Strategy. }
        \label{fig:res_uniform}
    \end{subfigure}
    \hfill
    \centering
    \begin{subfigure}[t]{0.24\textwidth}
        \centering
        \includegraphics[width=\linewidth]{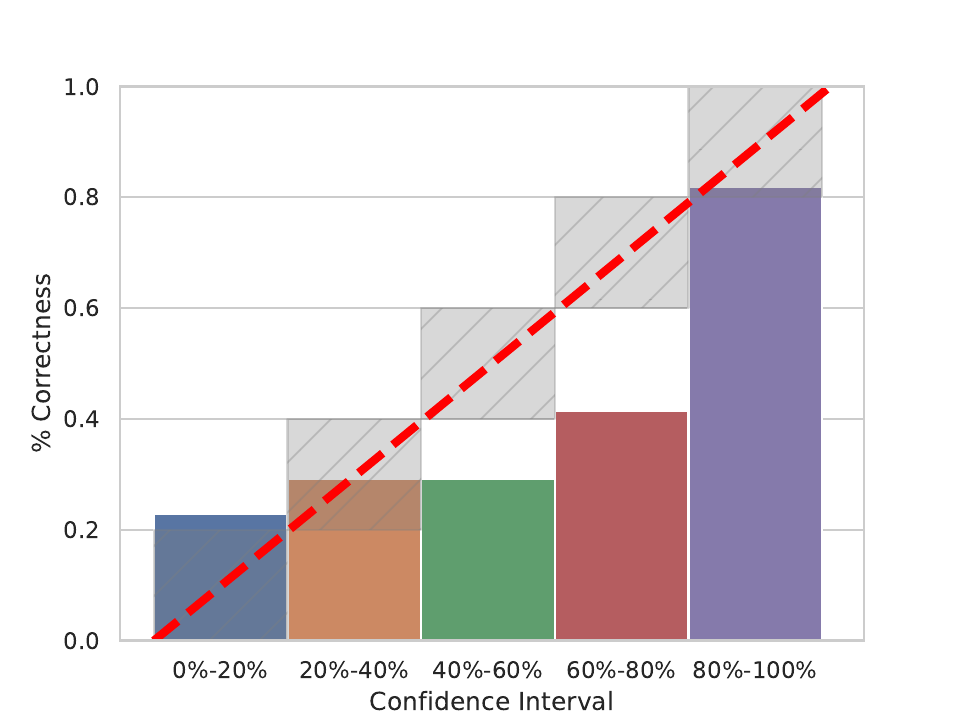}
        \caption{Confidence Estimation by Ignoring COE and Uniform Binning.}
        \label{fig:res_uniform_no_step1}
    \end{subfigure}\\
    \begin{subfigure}[t]{0.24\textwidth}
        \centering
        \includegraphics[width=\linewidth]{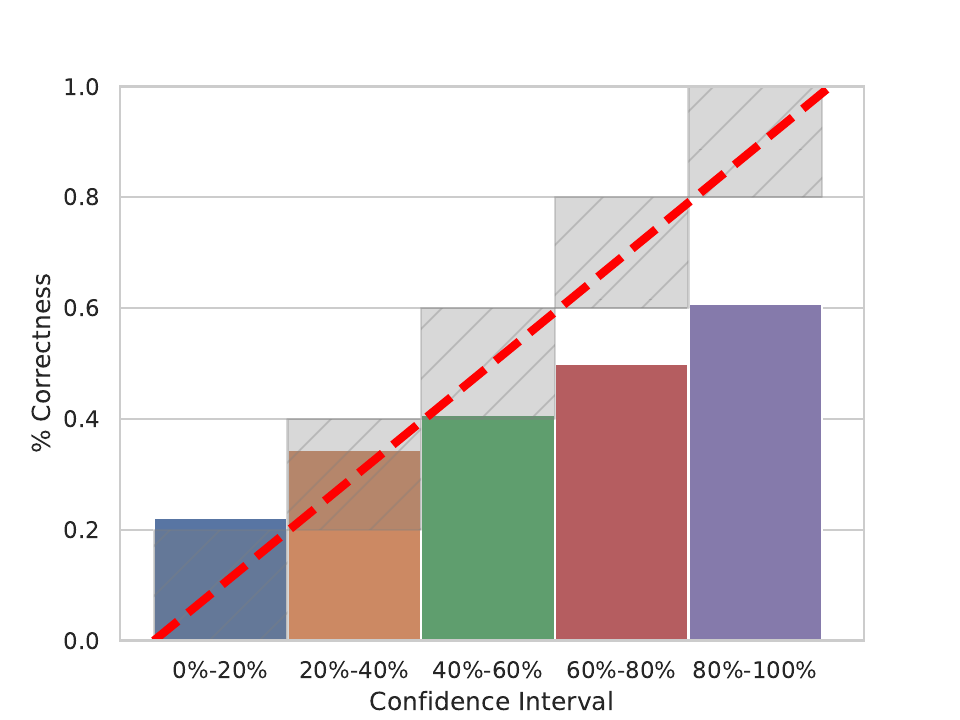}
        \caption{Confidence Estimation without Context and Calibrated Binning Strategy.}
        \label{fig:res_no_context}
        \end{subfigure}
        \hspace{3mm}
    \begin{subfigure}[t]{0.24\textwidth}
        \centering
        \includegraphics[width=\linewidth]{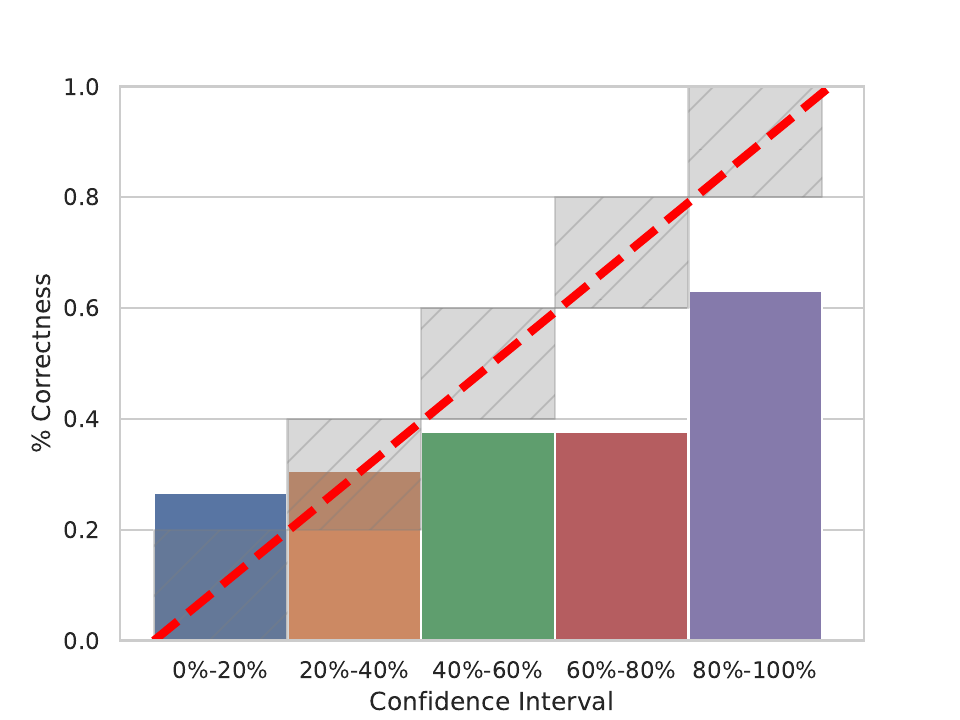}
        \caption{Confidence Estimation without Analysis and Calibrated Binning Strategy.}
\label{fig:res_one_step_with_context}
        \end{subfigure}
        \hspace{3mm}
    \begin{subfigure}[t]{0.24\textwidth}
        \centering
        \includegraphics[width=\linewidth]{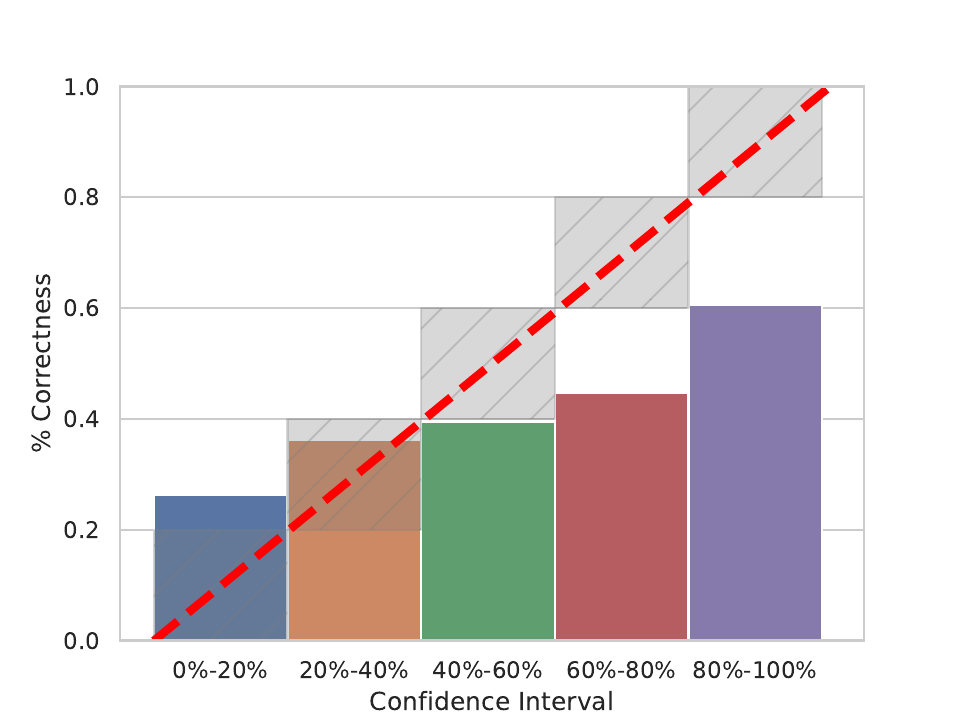}
        \caption{Confidence Estimation without Analysis \& without Context and Calibrated Binning Strategy.}
        \label{fig:res_one_step_wo_context}
    \end{subfigure}
    \caption{Reliability curve of calibrated confidence on root causes generated by GPT-4 with 8K tokens of context (including query incident).}
    \label{fig:reliability_gpt4}
\end{figure*}

Figure \ref{fig:res_uniform} shows the reliability diagram if we were to project scores linearly to the interval between 0 and 1 and Figure \ref{fig:res_uniform_no_step1} is the linearly projected RCE score. By comparing Figures \ref{fig:res_full} and \ref{fig:res_uniform}, we show that the optimization step for binning is necessary since neither model-produced score is scaled uniformly. 
\subsubsection{RQ2: \qtwo}
Our empirical study has provided compelling evidence of the significant advantages gained from incorporating historical incidents into the RCE scoring process. Without the introduction of this additional information, the RCE procedure would have lost its essential purpose and efficacy. Due to the highly domain-specific nature of the problem, it inherently relies on external sources of information that were not present in its original pre-training corpus. This stands in contrast to tasks involving commonsense question-answering or general reasoning in the NLP domain, where the necessary knowledge is already available to the model via pre-training.

In scenarios where the model must generate analyses and judgments about root causes without the context of historical incidents, it tends to lean on general-domain knowledge and superficial correspondences between incident descriptions and potential root causes. Such reliance is often insufficient, because the model, without historical context or service-specific insights, tends to fail to grasp the nuanced, context-specific elements and foundational patterns essential in thoroughly examining predicted root causes. Incident descriptions alone do not encapsulate the root cause, leading to a lack of depth and precision in the model's analysis. Consequently, in the absence of historical incidents and service-specific knowledge, the accurate evaluation of a root cause proves impossible.

By contrast, in scenarios where historical incidents are provided, the model becomes capable of analyzing the similarities between the current incident and past occurrences, allowing it to compare the model-generated root causes with the actual root causes of these historical incidents. 
\begin{figure}[!htb]
  \centering
\begin{tcolorbox}[colback=blue!10,colframe=blue!50!black,title=A High-Confidence Prediction]
\textbf{Model Generated Root Cause}

Physical ARM64 machines experience occasional failures due to lost heartbeat or bad updates, requiring intervention to recover. Transitioning to ARM64 VMs could help reduce downtime and improve the pool's capacity.

\textbf{Ground Truth Root Cause}

Physical ARM64 machines sometimes lose heartbeat on reboot or are affected by a bad update and require intervention to recover. If possible, transition to ARM64 VMs could reduce our time spent recovering physical machines.

\end{tcolorbox}
  \caption{An Example of model-generated root cause with high confidence. The model predicted root cause indeed is correct.}
  \label{fig:correct_example}
\end{figure}
The integration of historical data not only enhances its understanding of complex, domain-specific nuances. With access to a rich database of past incidents, the model becomes adept at recognizing patterns and identifying recurring issues. In cases where a recurring issue is detected, the confidence estimator is empowered to directly verify the accuracy of the identified root cause. 

\subsubsection{RQ3: \qthree}

The emerging evidence surrounding chain-of-thought reasoning presents a promising avenue for bolstering the capabilities of a various spectrum of natural language processing (NLP) reasoning tasks. In our pursuit of harnessing this potential, we set our sights on investigating whether this approach could be seamlessly integrated into the scoring process, which is inherently characterized by subjectivity and open-endedness. Upon conducting our investigations, we observed that the language model does, in fact, derive substantial benefits from its earlier analyses which serves as a well-informed basis for it to make more reasonable and accurate scoring. 

\begin{figure}
    \centering
    \begin{subfigure}{0.3\textwidth}
        \centering
        \includegraphics[width=\linewidth]{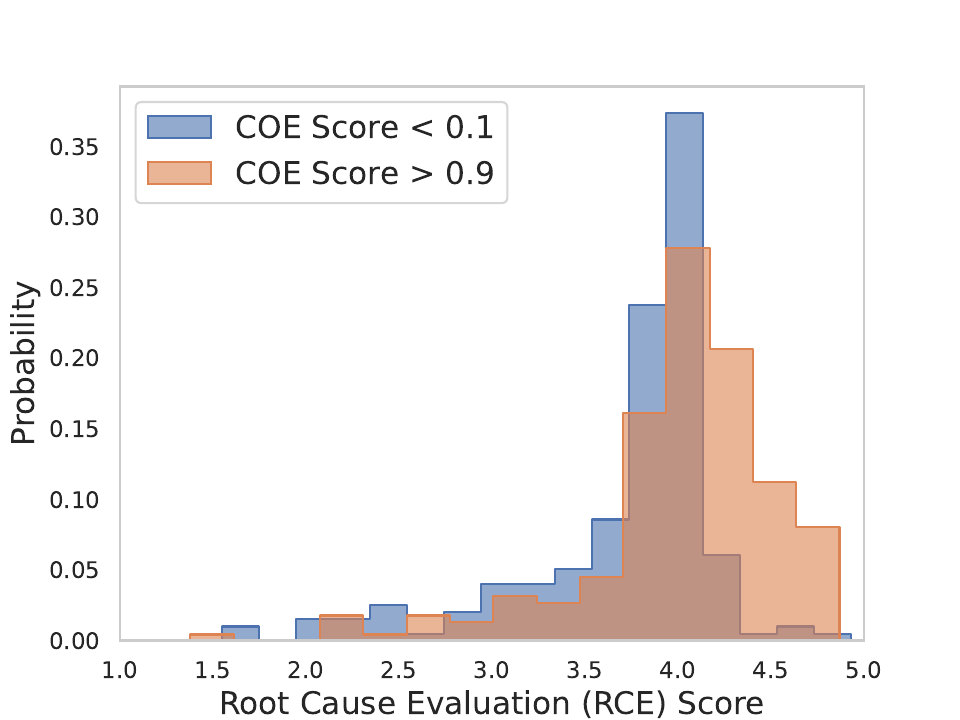}
        \caption{High ground-truth score data points.}
        \label{fig:dis_high}
    \end{subfigure}
    \hspace{4mm}
    \begin{subfigure}{0.3\textwidth}
        \centering
        \includegraphics[width=\linewidth]{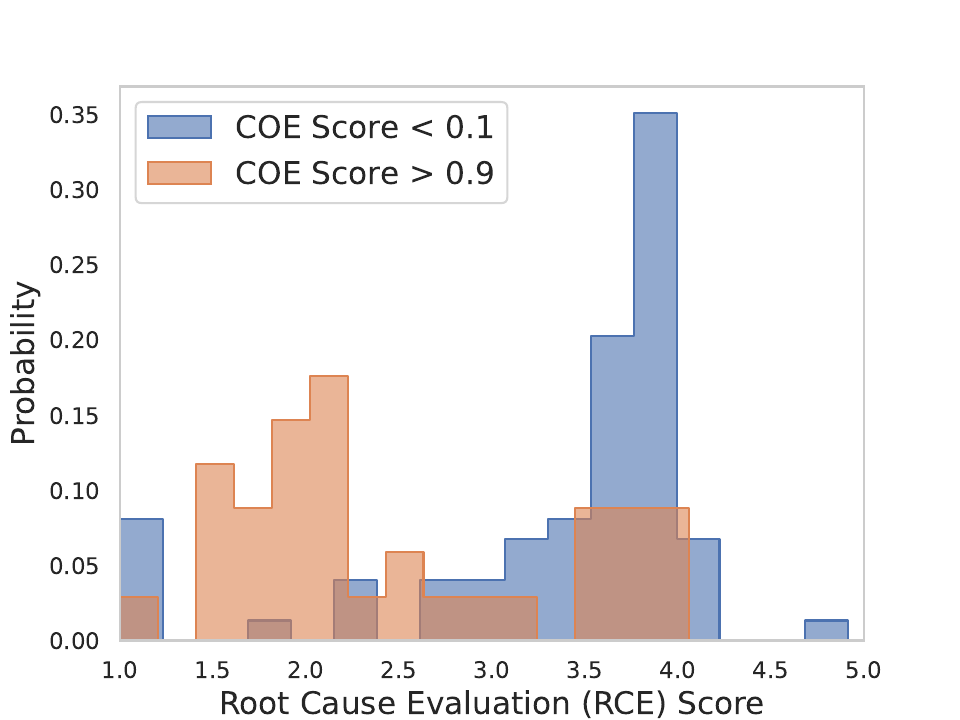}
        \caption{Low ground-truth score data points.}
        \label{fig:dis_low}
    \end{subfigure}

    \caption{Distribution of RCE scores. }
    \label{fig:rce_distribution}
\end{figure}

\begin{figure}
    \centering
    \begin{subfigure}[h]{0.3\textwidth}
        \centering
        \includegraphics[width=\linewidth]{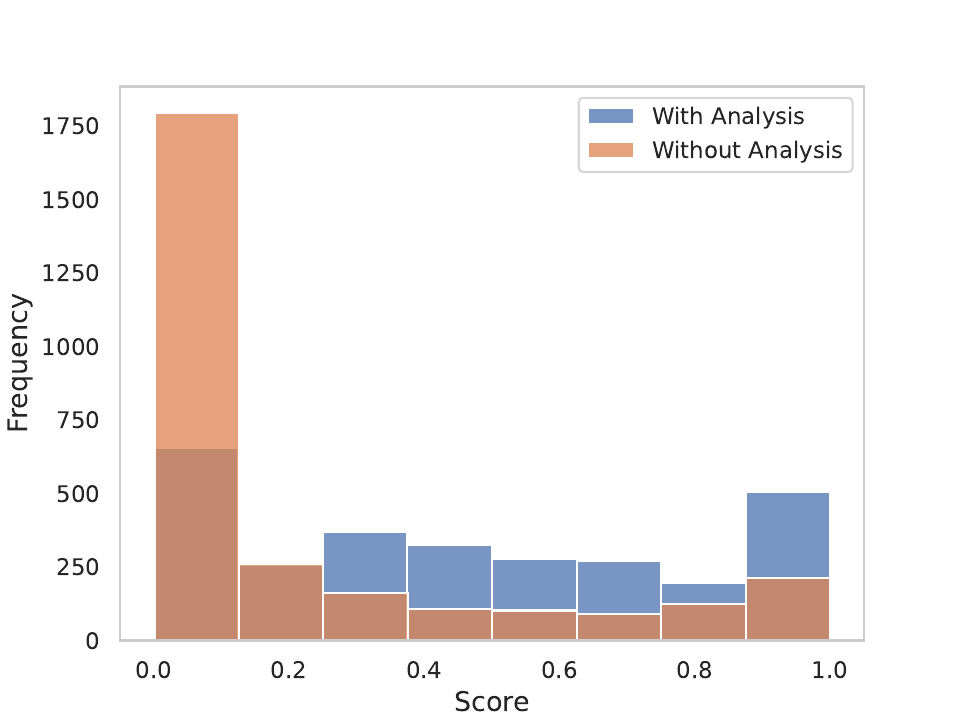}
        \caption{Distribution of COE scores with/without analysis.}
        \label{fig:dis}
    \end{subfigure}
    \hspace{4mm}
    \begin{subfigure}[h]{0.3\textwidth}
        \centering
        \includegraphics[width=\linewidth]{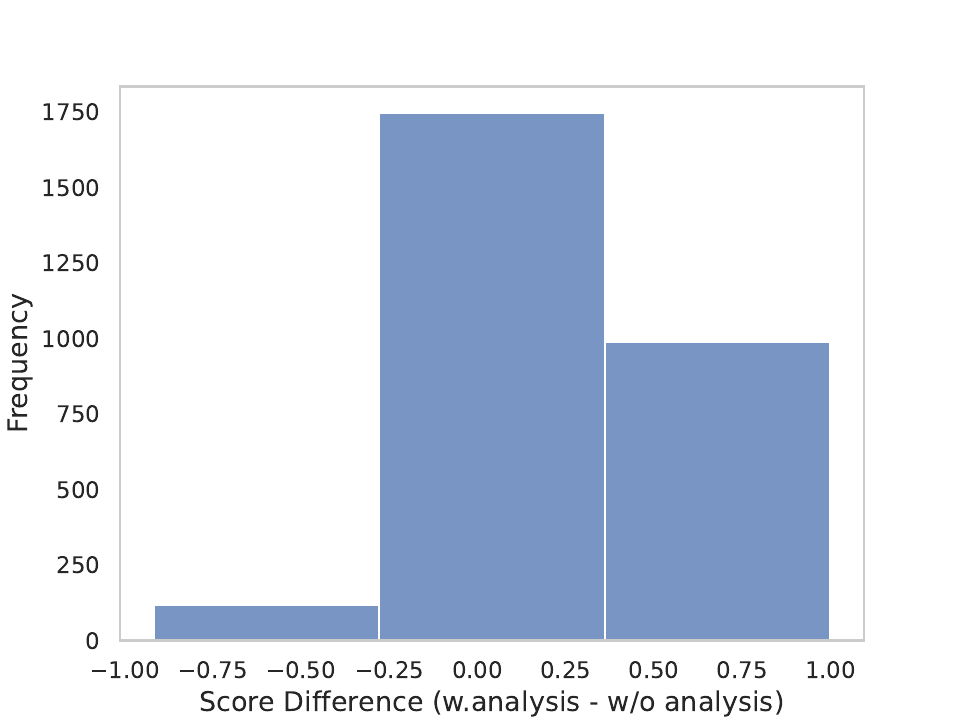}
        \caption{Distribution COE score difference w./w.o. analysis for each incident.}
        \label{fig:dis_low}
    \end{subfigure}

    \caption{COE scores. Lack of analysis makes the model unable to leverage its analysis to score rationally and therefore yields meaninglessly low scores. }
    \label{fig:coe_distribution}
\end{figure}

Notably, for COE scoring, direct scoring yields low scores for the majority of the cases since the question is highly non-trivial for the model to answer directly, thus the model tends to be confounded and biased towards generating low scores regardless of how informed it actually is, as shown in Figure~\ref{fig:coe_distribution}. For RCE, zero-shot scoring tends to perform worse in separating the correct and incorrect cases (Figure ~\ref{fig:rce_distribution}), thus yielding unsatisfactory confidence estimation results.

The empirical results imply that generating analysis is an important component of our approach in this task, as it empowers the model to provide comprehensive insights while addressing the specific questions outlined in the instructions. By generating analysis, the language model can delve deeper into the given context, extracting relevant information, and offering well-structured responses that cater precisely to the task at hand. Through this process, the model can identify key examples within the retrieved context that hold particular significance in making informed judgments. By highlighting these crucial examples, the model enhances its ability to prioritize relevant information for the RCE step, ultimately leading to more accurate responses.







\subsubsection{RQ4: \qfour}
\begin{figure*}[!htb]
    \centering
        \begin{subfigure}[h]{0.23\textwidth}
        \centering
        \includegraphics[width=\linewidth]{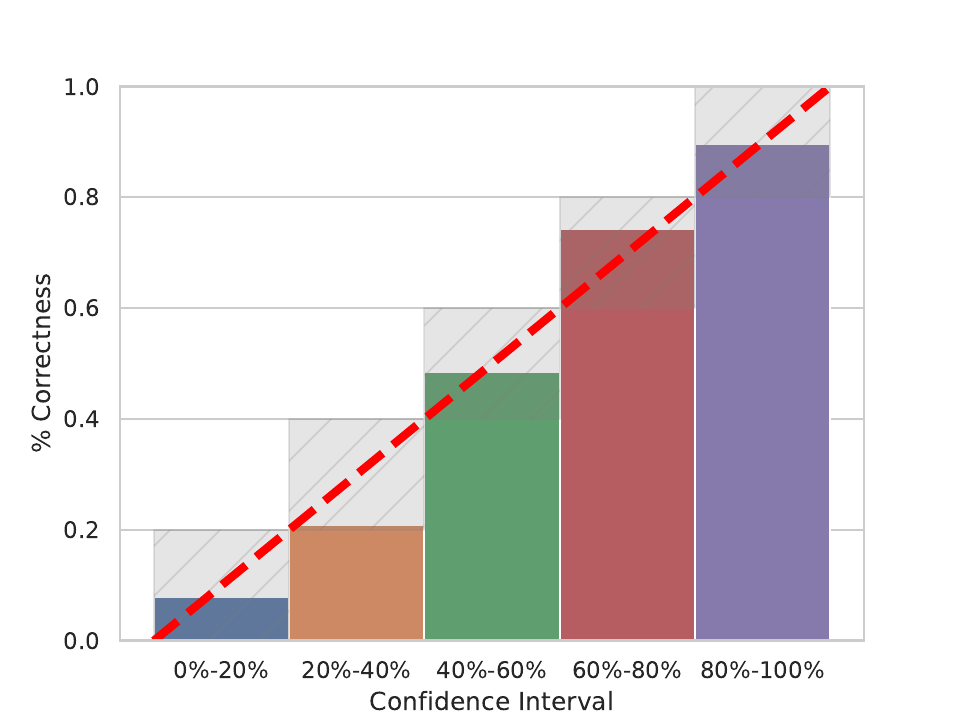}
        \caption{Calibration Result for Text-DaVinci-003-3shot.}
        \label{fig:davinci3shotfull}
    \end{subfigure}
    \hfill
    \centering
    \begin{subfigure}[h]{0.23\textwidth}
        \centering
        \includegraphics[width=\linewidth]{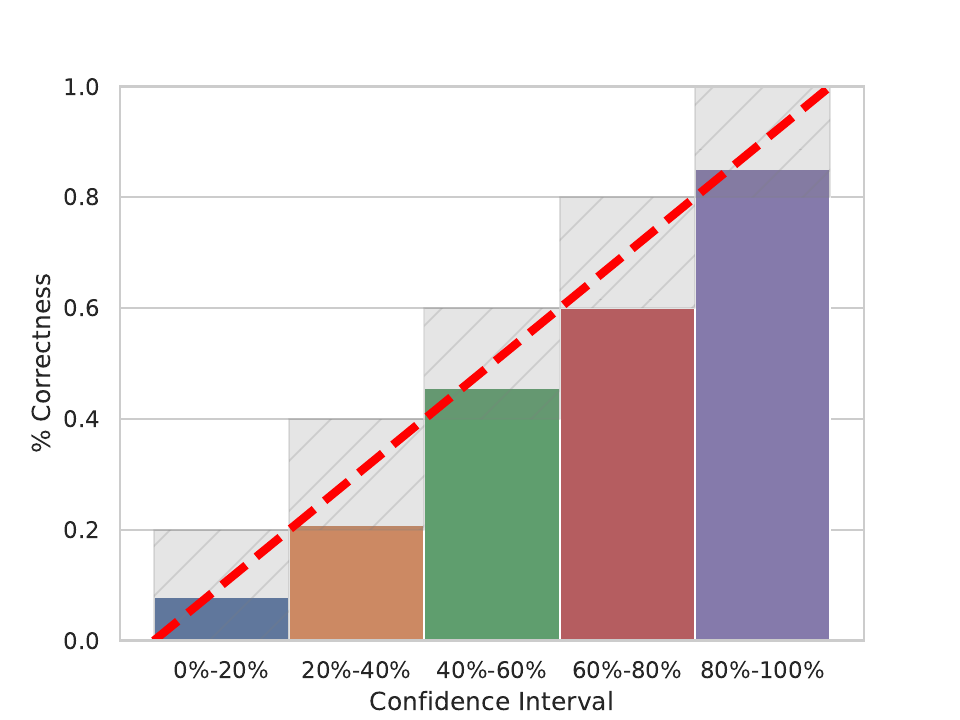}
        \caption{Calibration Result for GPT-3.5 Turbo-3shot.}
        \label{fig:chat3shotfull}
    \end{subfigure}
    \hfill
    \begin{subfigure}[h]{0.23\textwidth}
        \centering
        \includegraphics[width=\linewidth]{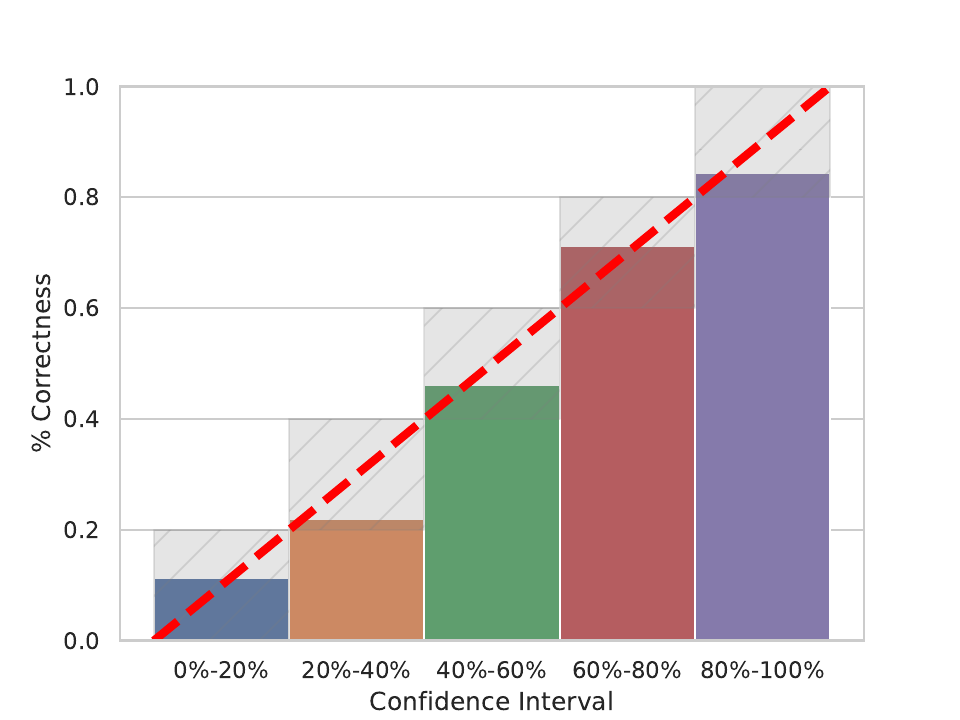}
        \caption{Calibration Result for GPT-3.5-Turbo-3896.}
        \label{fig:chat4kfull}
    \end{subfigure}
        \hfill
    \begin{subfigure}[h]{0.23\textwidth}
        \centering
        \includegraphics[width=\linewidth]{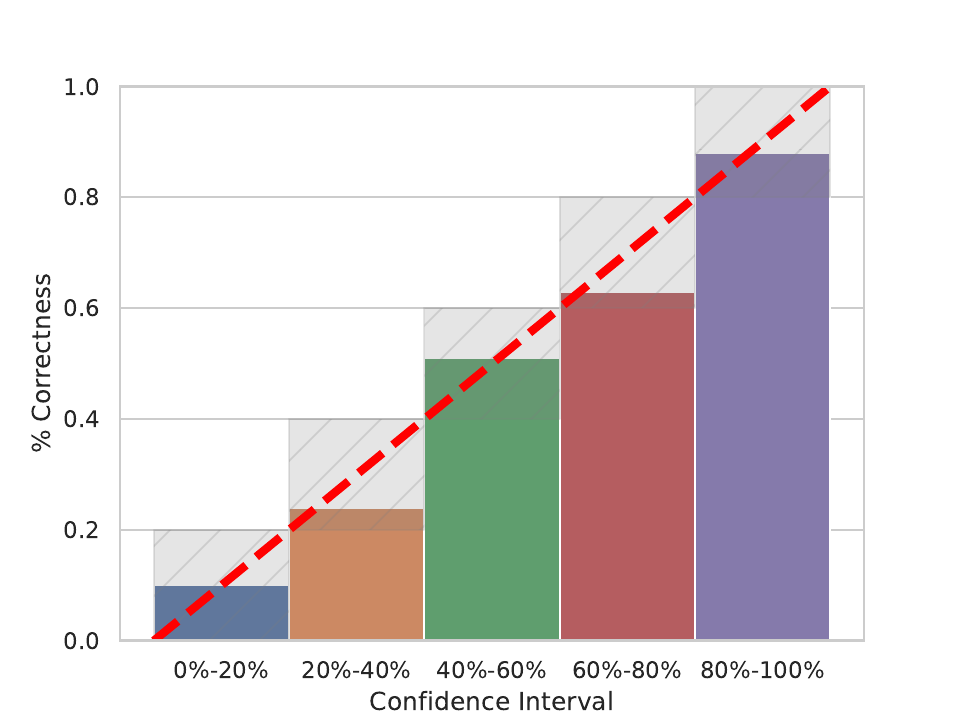}
        \caption{Calibration Result for Text-DaVinci-003-3896.}
        \label{fig:davinci4kfull}
    \end{subfigure}
    \\
    \begin{subfigure}[h]{0.23\textwidth}
        \centering
        \includegraphics[width=\linewidth]{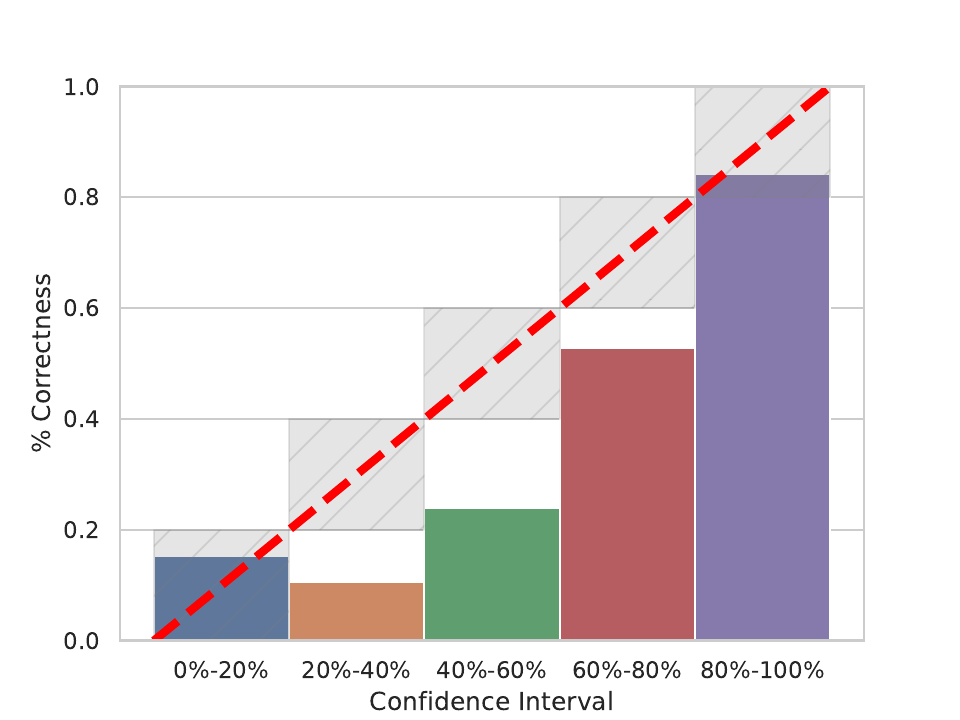}
        \caption{Calibration Result for Text-DaVinci-003-3shot w/o COE.}
        \label{fig:davinci3shot_step2}
    \end{subfigure}
    \hfill
    \centering
    \begin{subfigure}[h]{0.23\textwidth}
        \centering
        \includegraphics[width=\linewidth]{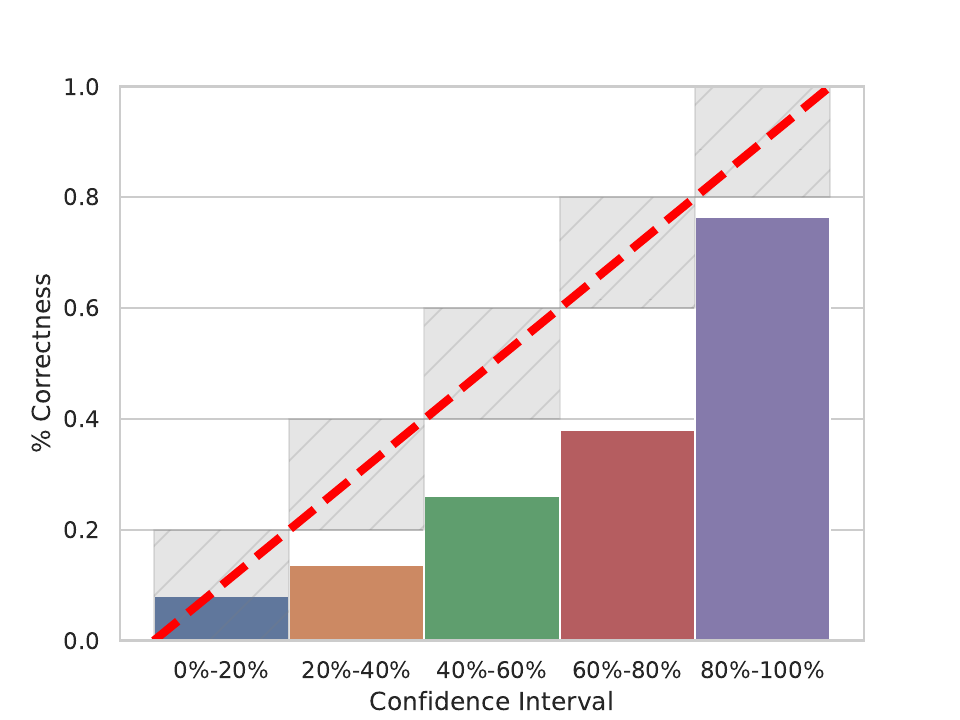}
        \caption{Calibration Result for GPT-3.5-Turbo-3896 w/o COE.}
        \label{fig:chat3shot_step2}
    \end{subfigure}
    \hfill
    \begin{subfigure}[h]{0.23\textwidth}
        \centering
        \includegraphics[width=\linewidth]{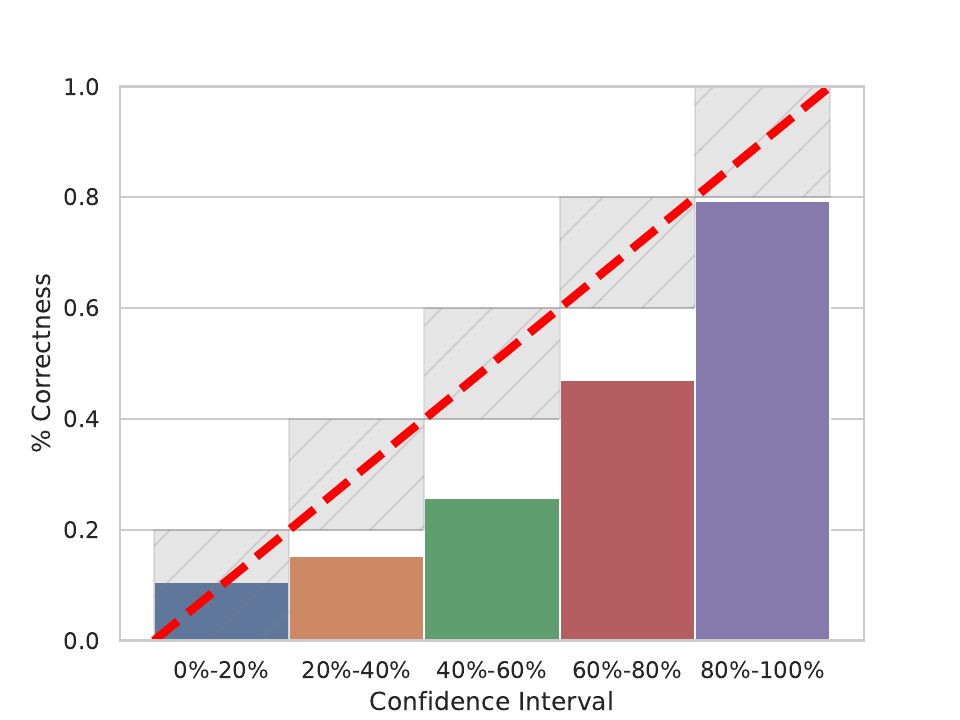}
        \caption{Calibration Result for GPT-3.5-Turbo-3896 w/o COE.}
        \label{fig:chat4k_step2}
    \end{subfigure}
    \hfill
        \begin{subfigure}[h]{0.23\textwidth}
        \centering
        \includegraphics[width=\linewidth]{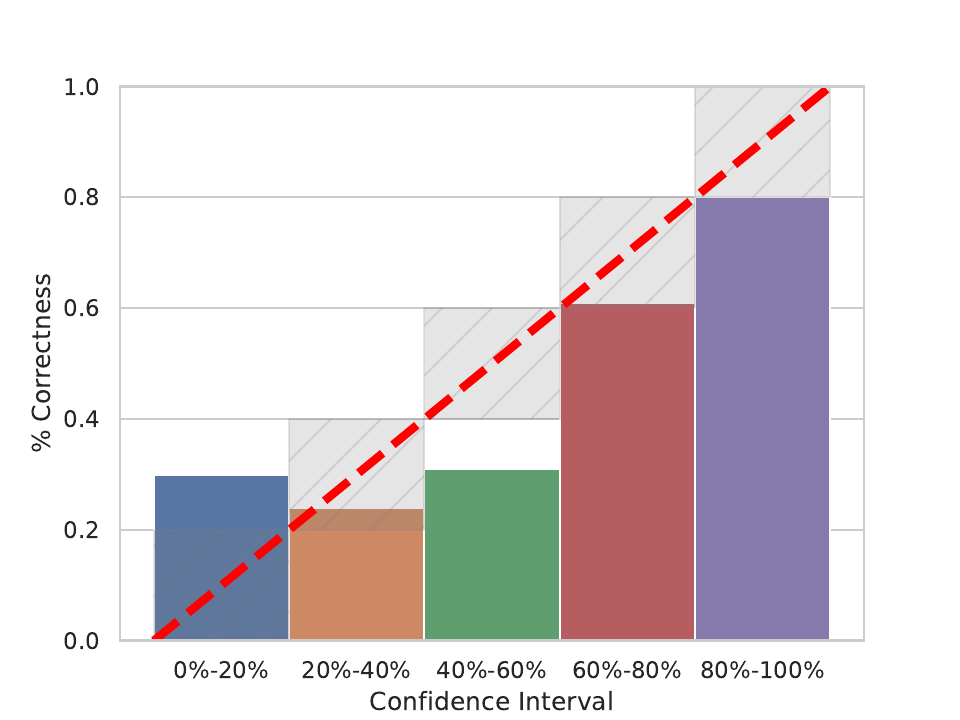}
        \caption{Calibration Result for Text-DaVinci-003-3896 w/o COE.}
        \label{fig:davinci4k_step2}
    \end{subfigure}
    \caption{Reliability Diagrams of GPT-3.5-Turbo and Text-DaVinci-003 Experiments. Our approach produces accurately calibrated scores in these settings.}
    \label{fig:reliability_other}
\end{figure*}

\begin{table*}[!htp]\centering
\begin{center}
\small
\begin{tabular}{lccccc}\toprule
\textbf{\shortstack{Root Cause Generator}} & \textbf{\toolname} & \textbf{\shortstack{RCE-Only\\With Optimization}} & \textbf{\shortstack{COE and RCE\\Uniform Binning}} & \textbf{\shortstack{RCE-Only\\Uniform Binning}} \\\midrule
GPT-4-8K &0.082 &0.157 \textcolor{red}{\textbf{(+0.075)}} &0.118 \textcolor{red}{\textbf{(+0.036)}} &0.261 \textcolor{red}{\textbf{(+0.179)}} \\
GPT-3.5-Turbo-3shot &0.084 &0.192 \textcolor{red}{\textbf{(+0.108)}} &0.121 \textcolor{red}{\textbf{(+0.037)}} &0.327 \textcolor{red}{\textbf{(+0.243)}} \\
Text-DaVinci-003-3shot &0.081 &0.219 \textcolor{red}{\textbf{(+0.138)}} &0.132 \textcolor{red}{\textbf{(+0.051)}} &0.459 \textcolor{red}{\textbf{(+0.378)}} \\
GPT-3.5-Turbo-3896 &0.053 &0.156 \textcolor{red}{\textbf{(+0.103)}} &0.112 \textcolor{red}{\textbf{(+0.059)}} &0.355 \textcolor{red}{\textbf{(+0.302)}} \\
Text-DaVinci-003-3896 &0.072 &0.189 \textcolor{red}{\textbf{(+0.117)}} &0.176 \textcolor{red}{\textbf{(+0.104)}} &0.395 \textcolor{red}{\textbf{(+0.323)}} \\
\bottomrule
\end{tabular}
\caption{ECE scores under different settings. Numbers in parentheses are ECE differences compared with the full method. Lower ECE scores indicate better calibration performance. Full method always perform the best in this table.}\label{tab:main}
\end{center}
\end{table*}
Here we examine whether the proposed method can indeed work for root causes generated by different upstream root cause generators with various amounts of retrieved information, which implies varying accuracies for upstream root cause analysis results as well as the distributions of the underlying text.
\begin{figure}[htb]
    \centering
    \begin{subfigure}[h]{.5\textwidth}  
        \centering
        \includegraphics[width=\textwidth]{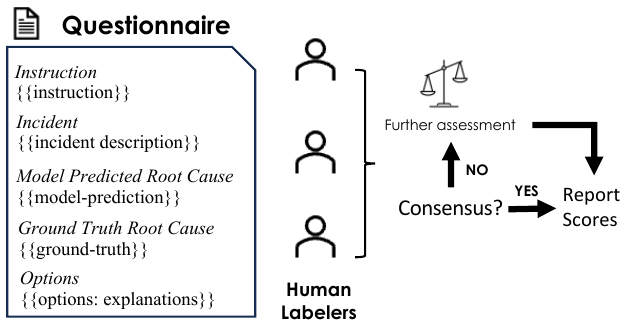}
        \caption{Human evaluation procedure with consensus labeling.}
        \label{fig:enter-label}
    \end{subfigure}
    \begin{subfigure}[h]{.4\textwidth}  
        \centering
        \includegraphics[width=\textwidth]{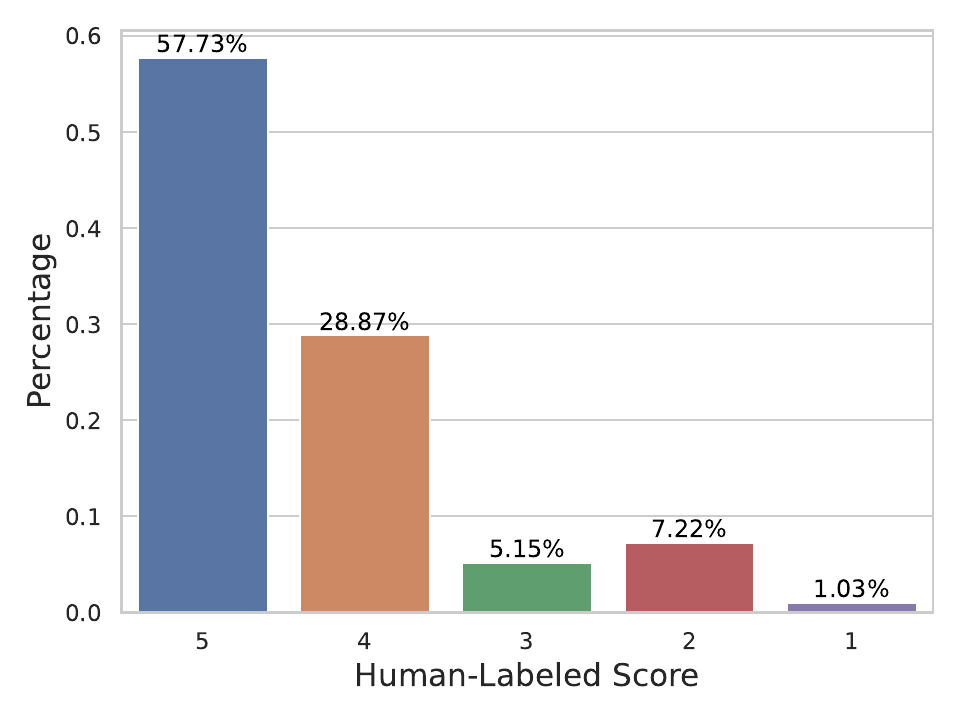}
        \caption{Human-evaluation results for high-confidence cases.}
        \label{fig:human-eval}
    \end{subfigure}
    \caption{Human Evaluation}  
    \label{fig:overall-label}  
\end{figure}
By conducting experiments using the underlying causes produced by retrieval-augmented Text-DaVinci-003 applied to the three most pertinent incidents (depicted in Figure \ref{fig:davinci3shotfull}), along with GPT-3.5-Turbo's responses to the top three relevant incidents (illustrated in Figure \ref{fig:chat3shotfull}), as well as in scenarios with maximum context length (as presented in Figure \ref{fig:chat4kfull}), we demonstrate the persistent effectiveness of the proposed approach. This applicability holds true across diverse output distributions from models both fine-tuned with human feedback (such as GPT-3.5-Turbo and GPT-4) and those that are not (like Text-DaVinci-003), as well as when exposed to varying amounts of information.

\subsection{Human Evaluation}
The human evaluation setup involves assessing the quality of model-generated root causes in comparison to the ground truth root causes. This evaluation process is performed by human evaluators who are tasked with assigning a similarity score to each model-generated root cause based on a predefined scale. The goal is to determine how well the model's outputs align with the actual root causes.
We present 5 options to human labelers (including the score and the corresponding tier):\\
\textbf{1} {\textit{(Totally Irrelevant)}};
\textbf{2} {\textit{(Vague or Distant Relevance)}};
\textbf{3} {\textit{(Has Similarities)}};
\textbf{4} {\textit{(Clearly Similar)}};
\textbf{5} {\textit{(Very Similar and Highly Relevant)}}.\\
We employ a consensus labeling approach, wherein an item receives a label if a consensus is achieved among the minimum required number of labelers. Should a consensus not be reached, the item undergoes further assessment by additional labelers. The minimum required number of labelers in our case is 3. 

In real-world cloud incident troubleshooting scenarios, the most critical instances are those where the model assigns high confidence scores to the predicted root causes. In such cases, on-call engineers are more inclined to trust these predictions, leading to significant consequences if the estimates prove inaccurate.

Therefore, we labeled the root causes generated from GPT-4 8K token falling into the 80\% - 100\% bin in the full method, the label distribution is shown in Figure \ref{fig:human-eval}. 
We notice that 86\% of the cases (scored 5 and 4) are labeled by humans as highly similar to ground truth root causes, which falls within the range of 80\% - 100\%. The human evaluation results further demonstrate that the calibration approach is indeed effective as it can produce results consistent with human judgment.







\section{Related Work}
\subsection{Automated Root Cause Analysis with Language Models}
Machine learning as applied to IT operations, often termed as MLOps or AIOps, has been studied in the context of cloud incident management for a while. For close to a decade, various studies have underscored its effectiveness in managing and resolving incidents in cloud services\cite{chen2019onlinesys,chen2019triage,azad2021picking,nair2015learning,bansal2020decaf,luo2014timeseries,jiang2020deeprecommand}. The progression in this field has seen a myriad of approaches; from methodologies that leverage time-series and feature-based techniques to the ones employing advanced generative language models. The underlying motive has always been to harness the power of the colossal amount of data available, conceptualize effective machine-learning solutions, and thereby, augment human capabilities and enhance overall productivity.

In the recent past, we've witnessed a paradigm shift, particularly in the capabilities of language models. These models have not only evolved but also showcased proficiency in managing domain-specific question-answering tasks \cite{wang2023industrydomainqa}. The implications of such advancements are profound. They have carved out potential pathways where these models can be seamlessly integrated into the product's life cycle. A case in point is the root cause analysis (RCA) procedure which has also been studied by a series of works like \cite{ahmed2023rca,chen2023practicalrca,jin2023outage}. However, one common problem is that language models are knowingly good at hallucinating \cite{zhang2023snowball,ji2023hallucinatesurvey}, which can generate outputs that are seemingly plausible while containing highly misleading information that could confuse the stakeholders taking the model outputs as reference. 



\subsection{Confidence Calibration Of Pre-trained Language Models}

Confidence calibration for pre-trained language models has been studied under several question-answering set ups.
They explored various methods to calibrate pre-trained language models for QA tasks. Mielke et al. \cite{mielke2022reducing} proposed to train a separate confidence calibrator, but their approach requires representations from the models. Lin et al \cite{lin2022teachingtoexpress} proposed to use the verbalized confidences by the models as confidence estimates, but it requires fine-tuning the model that is used to produce answers to perform this function, which is not always feasible.

Confidence estimation under an in-context learning set-up is more challenging as it is more constrained in terms of the components we can manipulate. It has been pointed out by Zhou et al. \cite{zhou2023overconfidence}  that guiding models to express their uncertainties verbally via prompting could harm the model's performance. Zhao et al. \cite{zhao2023paretocalibration} 
proposed a calibration framework tailored for classification tasks, requiring finite output space and a hand-crafted supervision function, and therefore cannot be applied to open-ended question-answering problems.  Beyond the categorical set-up, recent works \cite{xiong2023express,tian2023askforcalibration} explored prompting-based confidence estimation for factual and numerical QA tasks. Xiong et al. \cite{xiong2023express} pointed out that their confidence estimation approaches struggle in more complex tasks particularly those requiring domain-specific knowledge. In particular, Jiang et al. \cite{jiang2021knowitknow} demonstrated that augmenting inputs with retrieved knowledge can benefit confidence estimation in question-answering. However, their study was conducted over either multiple-choice or extractive QA tasks, both of which are essentially classification problems without needing to generate open-ended responses.

\section{Conclusion}

In light of our exploration into the challenges of the rapidly evolving AIOps applications, particularly concerning cloud-based platforms, this paper identifies the major obstacles tied to the adoption of LLM-based root cause predictors. These include the potential for incorrect recommendations and undetectable hallucinations, issues that could misdirect on-call engineers and lead to substantial losses. We propose to perform confidence estimation for each prediction in order to mitigate these challenges.
We developed a general framework that makes minimal assumptions on the root cause predictor, \toolname, designed with the flexibility to encompass various root cause predictors. Our approach breaks the confidence estimation process into two phases, where we first measure the strength of evidence provided by historical references, then quantify the quality of the model prediction. We obtain final confidence assignments from the two scores with an optimization step.
We have demonstrated the adaptability and effectiveness of our approach across diverse output distributions. Notably, this research stands as a pioneering effort in prompting retrieval-augmented LLMs for confidence calibration in open-ended question-answering tasks in domain-specific scenarios. More broadly, this work also serves as evidence to prove the ability of LLMs to be prompted to produce calibrated confidence estimates. It also suggests their potential to be used for confidence estimation in various other domain-specific tasks to foster more reliable AIOps systems. 
\begin{acks}
We thank Yangyi Chen, Chi Han, Jialiang Xu, and Liliang Ren for their constructive discussions and valuable feedback on this work. 
\end{acks}

\bibliographystyle{ACM-Reference-Format}
\bibliography{sample-base}


\begin{thebibliography}{37}


\ifx \showCODEN    \undefined \def \showCODEN     #1{\unskip}     \fi
\ifx \showDOI      \undefined \def \showDOI       #1{#1}\fi
\ifx \showISBNx    \undefined \def \showISBNx     #1{\unskip}     \fi
\ifx \showISBNxiii \undefined \def \showISBNxiii  #1{\unskip}     \fi
\ifx \showISSN     \undefined \def \showISSN      #1{\unskip}     \fi
\ifx \showLCCN     \undefined \def \showLCCN      #1{\unskip}     \fi
\ifx \shownote     \undefined \def \shownote      #1{#1}          \fi
\ifx \showarticletitle \undefined \def \showarticletitle #1{#1}   \fi
\ifx \showURL      \undefined \def \showURL       {\relax}        \fi
\providecommand\bibfield[2]{#2}
\providecommand\bibinfo[2]{#2}
\providecommand\natexlab[1]{#1}
\providecommand\showeprint[2][]{arXiv:#2}

\bibitem[Ahmed et~al\mbox{.}(2023)]%
        {ahmed2023rca}
\bibfield{author}{\bibinfo{person}{Toufique Ahmed}, \bibinfo{person}{Supriyo Ghosh}, \bibinfo{person}{Chetan Bansal}, \bibinfo{person}{Thomas Zimmermann}, \bibinfo{person}{Xuchao Zhang}, {and} \bibinfo{person}{Saravan Rajmohan}.} \bibinfo{year}{2023}\natexlab{}.
\newblock \bibinfo{title}{Recommending Root-Cause and Mitigation Steps for Cloud Incidents using Large Language Models}.
\newblock
\newblock
\showeprint[arxiv]{2301.03797}~[cs.SE]


\bibitem[Azad et~al\mbox{.}(2021)]%
        {azad2021picking}
\bibfield{author}{\bibinfo{person}{Amar~Prakash Azad}, \bibinfo{person}{Supriyo Ghosh}, \bibinfo{person}{Ajay Gupta}, \bibinfo{person}{Harshit Kumar}, {and} \bibinfo{person}{Prateeti Mohapatra}.} \bibinfo{year}{2021}\natexlab{}.
\newblock \bibinfo{title}{Picking Pearl From Seabed: Extracting Artefacts from Noisy Issue Triaging Collaborative Conversations for Hybrid Cloud Services}.
\newblock
\newblock
\showeprint[arxiv]{2105.15065}~[cs.AI]


\bibitem[Bang et~al\mbox{.}(2023)]%
        {bang2023multitask}
\bibfield{author}{\bibinfo{person}{Yejin Bang}, \bibinfo{person}{Samuel Cahyawijaya}, \bibinfo{person}{Nayeon Lee}, \bibinfo{person}{Wenliang Dai}, \bibinfo{person}{Dan Su}, \bibinfo{person}{Bryan Wilie}, \bibinfo{person}{Holy Lovenia}, \bibinfo{person}{Ziwei Ji}, \bibinfo{person}{Tiezheng Yu}, \bibinfo{person}{Willy Chung}, {et~al\mbox{.}}} \bibinfo{year}{2023}\natexlab{}.
\newblock \showarticletitle{A multitask, multilingual, multimodal evaluation of chatgpt on reasoning, hallucination, and interactivity}.
\newblock \bibinfo{journal}{\emph{arXiv preprint arXiv:2302.04023}} (\bibinfo{year}{2023}).
\newblock


\bibitem[Bansal et~al\mbox{.}(2020)]%
        {bansal2020decaf}
\bibfield{author}{\bibinfo{person}{Chetan Bansal}, \bibinfo{person}{Sundararajan Renganathan}, \bibinfo{person}{Ashima Asudani}, \bibinfo{person}{Olivier Midy}, {and} \bibinfo{person}{Mathru Janakiraman}.} \bibinfo{year}{2020}\natexlab{}.
\newblock \showarticletitle{{DeCaf}}. In \bibinfo{booktitle}{\emph{Proceedings of the {ACM}/{IEEE} 42nd International Conference on Software Engineering: Software Engineering in Practice}}. \bibinfo{publisher}{{ACM}}.
\newblock
\urldef\tempurl%
\url{https://doi.org/10.1145/3377813.3381353}
\showDOI{\tempurl}


\bibitem[Bubeck et~al\mbox{.}(2023)]%
        {bubeck2023sparks}
\bibfield{author}{\bibinfo{person}{Sébastien Bubeck}, \bibinfo{person}{Varun Chandrasekaran}, \bibinfo{person}{Ronen Eldan}, \bibinfo{person}{Johannes Gehrke}, \bibinfo{person}{Eric Horvitz}, \bibinfo{person}{Ece Kamar}, \bibinfo{person}{Peter Lee}, \bibinfo{person}{Yin~Tat Lee}, \bibinfo{person}{Yuanzhi Li}, \bibinfo{person}{Scott Lundberg}, \bibinfo{person}{Harsha Nori}, \bibinfo{person}{Hamid Palangi}, \bibinfo{person}{Marco~Tulio Ribeiro}, {and} \bibinfo{person}{Yi Zhang}.} \bibinfo{year}{2023}\natexlab{}.
\newblock \bibinfo{title}{Sparks of Artificial General Intelligence: Early experiments with GPT-4}.
\newblock
\newblock
\showeprint[arxiv]{2303.12712}~[cs.CL]


\bibitem[Chen et~al\mbox{.}(2019a)]%
        {chen2019onlinesys}
\bibfield{author}{\bibinfo{person}{Junjie Chen}, \bibinfo{person}{Xiaoting He}, \bibinfo{person}{Qingwei Lin}, \bibinfo{person}{Yong Xu}, \bibinfo{person}{Hongyu Zhang}, \bibinfo{person}{Dan Hao}, \bibinfo{person}{Feng Gao}, \bibinfo{person}{Zhangwei Xu}, \bibinfo{person}{Yingnong Dang}, {and} \bibinfo{person}{Dongmei Zhang}.} \bibinfo{year}{2019}\natexlab{a}.
\newblock \showarticletitle{An empirical investigation of incident triage for online service systems}. In \bibinfo{booktitle}{\emph{2019 IEEE/ACM 41st International Conference on Software Engineering: Software Engineering in Practice (ICSE-SEIP)}}. IEEE, \bibinfo{pages}{111--120}.
\newblock


\bibitem[Chen et~al\mbox{.}(2019b)]%
        {chen2019triage}
\bibfield{author}{\bibinfo{person}{Junjie Chen}, \bibinfo{person}{Xiaoting He}, \bibinfo{person}{Qingwei Lin}, \bibinfo{person}{Hongyu Zhang}, \bibinfo{person}{Dan Hao}, \bibinfo{person}{Feng Gao}, \bibinfo{person}{Zhangwei Xu}, \bibinfo{person}{Yingnong Dang}, {and} \bibinfo{person}{Dongmei Zhang}.} \bibinfo{year}{2019}\natexlab{b}.
\newblock \showarticletitle{Continuous Incident Triage for Large-Scale Online Service Systems}. In \bibinfo{booktitle}{\emph{2019 34th IEEE/ACM International Conference on Automated Software Engineering (ASE)}}. \bibinfo{pages}{364--375}.
\newblock
\urldef\tempurl%
\url{https://doi.org/10.1109/ASE.2019.00042}
\showDOI{\tempurl}


\bibitem[Chen et~al\mbox{.}(2023)]%
        {chen2023practicalrca}
\bibfield{author}{\bibinfo{person}{Yinfang Chen}, \bibinfo{person}{Huaibing Xie}, \bibinfo{person}{Minghua Ma}, \bibinfo{person}{Yu Kang}, \bibinfo{person}{Xin Gao}, \bibinfo{person}{Liu Shi}, \bibinfo{person}{Yunjie Cao}, \bibinfo{person}{Xuedong Gao}, \bibinfo{person}{Hao Fan}, \bibinfo{person}{Ming Wen}, \bibinfo{person}{Jun Zeng}, \bibinfo{person}{Supriyo Ghosh}, \bibinfo{person}{Xuchao Zhang}, \bibinfo{person}{Chaoyun Zhang}, \bibinfo{person}{Qingwei Lin}, \bibinfo{person}{Saravan Rajmohan}, {and} \bibinfo{person}{Dongmei Zhang}.} \bibinfo{year}{2023}\natexlab{}.
\newblock \bibinfo{title}{Empowering Practical Root Cause Analysis by Large Language Models for Cloud Incidents}.
\newblock
\newblock
\showeprint[arxiv]{2305.15778}~[cs.SE]


\bibitem[Cosmides and Tooby(1996)]%
        {cosmides1996cognitive}
\bibfield{author}{\bibinfo{person}{Leda Cosmides} {and} \bibinfo{person}{John Tooby}.} \bibinfo{year}{1996}\natexlab{}.
\newblock \showarticletitle{Are Humans Good Intuitive Statisticians After All? Rethinking Some Conclusions from the Literature on Judgment Under Uncertainty}.
\newblock \bibinfo{journal}{\emph{Cognition}}  \bibinfo{volume}{58} (\bibinfo{date}{01} \bibinfo{year}{1996}), \bibinfo{pages}{1--73}.
\newblock
\urldef\tempurl%
\url{https://doi.org/10.1016/0010-0277(95)00664-8}
\showDOI{\tempurl}


\bibitem[Deng et~al\mbox{.}(2023)]%
        {deng2023large}
\bibfield{author}{\bibinfo{person}{Yinlin Deng}, \bibinfo{person}{Chunqiu~Steven Xia}, \bibinfo{person}{Chenyuan Yang}, \bibinfo{person}{Shizhuo~Dylan Zhang}, \bibinfo{person}{Shujing Yang}, {and} \bibinfo{person}{Lingming Zhang}.} \bibinfo{year}{2023}\natexlab{}.
\newblock \showarticletitle{Large language models are edge-case fuzzers: Testing deep learning libraries via fuzzgpt}.
\newblock \bibinfo{journal}{\emph{arXiv preprint arXiv:2304.02014}} (\bibinfo{year}{2023}).
\newblock


\bibitem[Dhuliawala et~al\mbox{.}(2023)]%
        {dhuliawala2023hallucinatechainofverification}
\bibfield{author}{\bibinfo{person}{Shehzaad Dhuliawala}, \bibinfo{person}{Mojtaba Komeili}, \bibinfo{person}{Jing Xu}, \bibinfo{person}{Roberta Raileanu}, \bibinfo{person}{Xian Li}, \bibinfo{person}{Asli Celikyilmaz}, {and} \bibinfo{person}{Jason Weston}.} \bibinfo{year}{2023}\natexlab{}.
\newblock \bibinfo{title}{Chain-of-Verification Reduces Hallucination in Large Language Models}.
\newblock
\newblock
\showeprint[arxiv]{2309.11495}~[cs.CL]


\bibitem[Guu et~al\mbox{.}(2020)]%
        {guu2020retrieval}
\bibfield{author}{\bibinfo{person}{Kelvin Guu}, \bibinfo{person}{Kenton Lee}, \bibinfo{person}{Zora Tung}, \bibinfo{person}{Panupong Pasupat}, {and} \bibinfo{person}{Mingwei Chang}.} \bibinfo{year}{2020}\natexlab{}.
\newblock \showarticletitle{Retrieval augmented language model pre-training}. In \bibinfo{booktitle}{\emph{International conference on machine learning}}. PMLR, \bibinfo{pages}{3929--3938}.
\newblock


\bibitem[Izacard et~al\mbox{.}(2022)]%
        {izacard2022retrievallm}
\bibfield{author}{\bibinfo{person}{Gautier Izacard}, \bibinfo{person}{Patrick Lewis}, \bibinfo{person}{Maria Lomeli}, \bibinfo{person}{Lucas Hosseini}, \bibinfo{person}{Fabio Petroni}, \bibinfo{person}{Timo Schick}, \bibinfo{person}{Jane Dwivedi-Yu}, \bibinfo{person}{Armand Joulin}, \bibinfo{person}{Sebastian Riedel}, {and} \bibinfo{person}{Edouard Grave}.} \bibinfo{year}{2022}\natexlab{}.
\newblock \showarticletitle{Few-shot learning with retrieval augmented language models}.
\newblock \bibinfo{journal}{\emph{arXiv preprint arXiv:2208.03299}} (\bibinfo{year}{2022}).
\newblock


\bibitem[Ji et~al\mbox{.}(2023)]%
        {ji2023hallucinatesurvey}
\bibfield{author}{\bibinfo{person}{Ziwei Ji}, \bibinfo{person}{Nayeon Lee}, \bibinfo{person}{Rita Frieske}, \bibinfo{person}{Tiezheng Yu}, \bibinfo{person}{Dan Su}, \bibinfo{person}{Yan Xu}, \bibinfo{person}{Etsuko Ishii}, \bibinfo{person}{Ye~Jin Bang}, \bibinfo{person}{Andrea Madotto}, {and} \bibinfo{person}{Pascale Fung}.} \bibinfo{year}{2023}\natexlab{}.
\newblock \showarticletitle{Survey of Hallucination in Natural Language Generation}.
\newblock \bibinfo{journal}{\emph{ACM Comput. Surv.}} \bibinfo{volume}{55}, \bibinfo{number}{12}, Article \bibinfo{articleno}{248} (\bibinfo{date}{mar} \bibinfo{year}{2023}), \bibinfo{numpages}{38}~pages.
\newblock
\showISSN{0360-0300}
\urldef\tempurl%
\url{https://doi.org/10.1145/3571730}
\showDOI{\tempurl}


\bibitem[Jiang et~al\mbox{.}(2020)]%
        {jiang2020deeprecommand}
\bibfield{author}{\bibinfo{person}{Jiajun Jiang}, \bibinfo{person}{Weihai Lu}, \bibinfo{person}{Junjie Chen}, \bibinfo{person}{Qingwei Lin}, \bibinfo{person}{Pu Zhao}, \bibinfo{person}{Yu Kang}, \bibinfo{person}{Hongyu Zhang}, \bibinfo{person}{Yingfei Xiong}, \bibinfo{person}{Feng Gao}, \bibinfo{person}{Zhangwei Xu}, \bibinfo{person}{Yingnong Dang}, {and} \bibinfo{person}{Dongmei Zhang}.} \bibinfo{year}{2020}\natexlab{}.
\newblock \showarticletitle{How to Mitigate the Incident? An Effective Troubleshooting Guide Recommendation Technique for Online Service Systems}. In \bibinfo{booktitle}{\emph{Proceedings of the 28th ACM Joint Meeting on European Software Engineering Conference and Symposium on the Foundations of Software Engineering}} (Virtual Event, USA) \emph{(\bibinfo{series}{ESEC/FSE 2020})}. \bibinfo{publisher}{Association for Computing Machinery}, \bibinfo{address}{New York, NY, USA}, \bibinfo{pages}{1410–1420}.
\newblock
\showISBNx{9781450370431}
\urldef\tempurl%
\url{https://doi.org/10.1145/3368089.3417054}
\showDOI{\tempurl}


\bibitem[Jiang et~al\mbox{.}(2021)]%
        {jiang2021knowitknow}
\bibfield{author}{\bibinfo{person}{Zhengbao Jiang}, \bibinfo{person}{Jun Araki}, \bibinfo{person}{Haibo Ding}, {and} \bibinfo{person}{Graham Neubig}.} \bibinfo{year}{2021}\natexlab{}.
\newblock \showarticletitle{How Can We Know When Language Models Know? On the Calibration of Language Models for Question Answering}.
\newblock \bibinfo{journal}{\emph{Transactions of the Association for Computational Linguistics (TACL)}}  \bibinfo{volume}{9} (\bibinfo{date}{9} \bibinfo{year}{2021}), \bibinfo{pages}{962--977}.
\newblock


\bibitem[Jin et~al\mbox{.}(2023)]%
        {jin2023outage}
\bibfield{author}{\bibinfo{person}{Pengxiang Jin}, \bibinfo{person}{Shenglin Zhang}, \bibinfo{person}{Minghua Ma}, \bibinfo{person}{Haozhe Li}, \bibinfo{person}{Yu Kang}, \bibinfo{person}{Liqun Li}, \bibinfo{person}{Yudong Liu}, \bibinfo{person}{Bo Qiao}, \bibinfo{person}{Chaoyun Zhang}, \bibinfo{person}{Pu Zhao}, \bibinfo{person}{Shilin He}, \bibinfo{person}{Federica Sarro}, \bibinfo{person}{Yingnong Dang}, \bibinfo{person}{Saravan Rajmohan}, \bibinfo{person}{Qingwei Lin}, {and} \bibinfo{person}{Dongmei Zhang}.} \bibinfo{year}{2023}\natexlab{}.
\newblock \bibinfo{title}{Assess and Summarize: Improve Outage Understanding with Large Language Models}.
\newblock
\newblock
\showeprint[arxiv]{2305.18084}~[cs.SE]


\bibitem[Li et~al\mbox{.}(2023)]%
        {li2023making}
\bibfield{author}{\bibinfo{person}{Yifei Li}, \bibinfo{person}{Zeqi Lin}, \bibinfo{person}{Shizhuo Zhang}, \bibinfo{person}{Qiang Fu}, \bibinfo{person}{Bei Chen}, \bibinfo{person}{Jian-Guang Lou}, {and} \bibinfo{person}{Weizhu Chen}.} \bibinfo{year}{2023}\natexlab{}.
\newblock \showarticletitle{Making Language Models Better Reasoners with Step-Aware Verifier}. In \bibinfo{booktitle}{\emph{Proceedings of the 61st Annual Meeting of the Association for Computational Linguistics (Volume 1: Long Papers)}}. \bibinfo{publisher}{Association for Computational Linguistics}, \bibinfo{address}{Toronto, Canada}, \bibinfo{pages}{5315--5333}.
\newblock
\urldef\tempurl%
\url{https://doi.org/10.18653/v1/2023.acl-long.291}
\showDOI{\tempurl}


\bibitem[Lin et~al\mbox{.}(2022)]%
        {lin2022teachingtoexpress}
\bibfield{author}{\bibinfo{person}{Stephanie Lin}, \bibinfo{person}{Jacob Hilton}, {and} \bibinfo{person}{Owain Evans}.} \bibinfo{year}{2022}\natexlab{}.
\newblock \bibinfo{title}{Teaching Models to Express Their Uncertainty in Words}.
\newblock
\newblock
\showeprint[arxiv]{2205.14334}~[cs.CL]


\bibitem[Luo et~al\mbox{.}(2014)]%
        {luo2014timeseries}
\bibfield{author}{\bibinfo{person}{Chen Luo}, \bibinfo{person}{Jian-Guang Lou}, \bibinfo{person}{Qingwei Lin}, \bibinfo{person}{Qiang Fu}, \bibinfo{person}{Rui Ding}, \bibinfo{person}{Dongmei Zhang}, {and} \bibinfo{person}{Zhe Wang}.} \bibinfo{year}{2014}\natexlab{}.
\newblock \showarticletitle{Correlating Events with Time Series for Incident Diagnosis}. In \bibinfo{booktitle}{\emph{Proceedings of the 20th ACM SIGKDD International Conference on Knowledge Discovery and Data Mining}} (New York, New York, USA) \emph{(\bibinfo{series}{KDD '14})}. \bibinfo{publisher}{Association for Computing Machinery}, \bibinfo{address}{New York, NY, USA}, \bibinfo{pages}{1583–1592}.
\newblock
\showISBNx{9781450329569}
\urldef\tempurl%
\url{https://doi.org/10.1145/2623330.2623374}
\showDOI{\tempurl}


\bibitem[Mialon et~al\mbox{.}(2023)]%
        {mialon2023augmentedsurvey}
\bibfield{author}{\bibinfo{person}{Gr{\'e}goire Mialon}, \bibinfo{person}{Roberto Dessi}, \bibinfo{person}{Maria Lomeli}, \bibinfo{person}{Christoforos Nalmpantis}, \bibinfo{person}{Ramakanth Pasunuru}, \bibinfo{person}{Roberta Raileanu}, \bibinfo{person}{Baptiste Roziere}, \bibinfo{person}{Timo Schick}, \bibinfo{person}{Jane Dwivedi-Yu}, \bibinfo{person}{Asli Celikyilmaz}, \bibinfo{person}{Edouard Grave}, \bibinfo{person}{Yann LeCun}, {and} \bibinfo{person}{Thomas Scialom}.} \bibinfo{year}{2023}\natexlab{}.
\newblock \showarticletitle{Augmented Language Models: a Survey}.
\newblock \bibinfo{journal}{\emph{Transactions on Machine Learning Research}} (\bibinfo{year}{2023}).
\newblock
\showISSN{2835-8856}
\urldef\tempurl%
\url{https://openreview.net/forum?id=jh7wH2AzKK}
\showURL{%
\tempurl}
\newblock
\shownote{Survey Certification}.


\bibitem[Mielke et~al\mbox{.}(2022)]%
        {mielke2022reducing}
\bibfield{author}{\bibinfo{person}{Sabrina~J. Mielke}, \bibinfo{person}{Arthur Szlam}, \bibinfo{person}{Emily Dinan}, {and} \bibinfo{person}{Y-Lan Boureau}.} \bibinfo{year}{2022}\natexlab{}.
\newblock \showarticletitle{Reducing Conversational Agents{'} Overconfidence Through Linguistic Calibration}.
\newblock \bibinfo{journal}{\emph{Transactions of the Association for Computational Linguistics}}  \bibinfo{volume}{10} (\bibinfo{year}{2022}), \bibinfo{pages}{857--872}.
\newblock
\urldef\tempurl%
\url{https://doi.org/10.1162/tacl_a_00494}
\showDOI{\tempurl}


\bibitem[Nair et~al\mbox{.}(2015)]%
        {nair2015learning}
\bibfield{author}{\bibinfo{person}{Vinod Nair}, \bibinfo{person}{Ameya Raul}, \bibinfo{person}{Shwetabh Khanduja}, \bibinfo{person}{Vikas Bahirwani}, \bibinfo{person}{Qihong Shao}, \bibinfo{person}{Sundararajan Sellamanickam}, \bibinfo{person}{Sathiya Keerthi}, \bibinfo{person}{Steve Herbert}, {and} \bibinfo{person}{Sudheer Dhulipalla}.} \bibinfo{year}{2015}\natexlab{}.
\newblock \showarticletitle{Learning a hierarchical monitoring system for detecting and diagnosing service issues}. In \bibinfo{booktitle}{\emph{Proceedings of the 21th ACM SIGKDD international conference on knowledge discovery and data mining}}. \bibinfo{pages}{2029--2038}.
\newblock


\bibitem[Nye et~al\mbox{.}(2021)]%
        {nye2021scratchpad}
\bibfield{author}{\bibinfo{person}{Maxwell Nye}, \bibinfo{person}{Anders~Johan Andreassen}, \bibinfo{person}{Guy Gur-Ari}, \bibinfo{person}{Henryk Michalewski}, \bibinfo{person}{Jacob Austin}, \bibinfo{person}{David Bieber}, \bibinfo{person}{David Dohan}, \bibinfo{person}{Aitor Lewkowycz}, \bibinfo{person}{Maarten Bosma}, \bibinfo{person}{David Luan}, {et~al\mbox{.}}} \bibinfo{year}{2021}\natexlab{}.
\newblock \showarticletitle{Show your work: Scratchpads for intermediate computation with language models}.
\newblock \bibinfo{journal}{\emph{arXiv preprint arXiv:2112.00114}} (\bibinfo{year}{2021}).
\newblock


\bibitem[Ouyang et~al\mbox{.}(2022)]%
        {ouyang2022instructions}
\bibfield{author}{\bibinfo{person}{Long Ouyang}, \bibinfo{person}{Jeffrey Wu}, \bibinfo{person}{Xu Jiang}, \bibinfo{person}{Diogo Almeida}, \bibinfo{person}{Carroll Wainwright}, \bibinfo{person}{Pamela Mishkin}, \bibinfo{person}{Chong Zhang}, \bibinfo{person}{Sandhini Agarwal}, \bibinfo{person}{Katarina Slama}, \bibinfo{person}{Alex Ray}, {et~al\mbox{.}}} \bibinfo{year}{2022}\natexlab{}.
\newblock \showarticletitle{Training language models to follow instructions with human feedback}.
\newblock \bibinfo{journal}{\emph{Advances in Neural Information Processing Systems}}  \bibinfo{volume}{35} (\bibinfo{year}{2022}), \bibinfo{pages}{27730--27744}.
\newblock


\bibitem[Rawte et~al\mbox{.}(2023)]%
        {rawte2023surveyhallucinate}
\bibfield{author}{\bibinfo{person}{Vipula Rawte}, \bibinfo{person}{Amit Sheth}, {and} \bibinfo{person}{Amitava Das}.} \bibinfo{year}{2023}\natexlab{}.
\newblock \showarticletitle{A Survey of Hallucination in Large Foundation Models}.
\newblock \bibinfo{journal}{\emph{arXiv preprint arXiv:2309.05922}} (\bibinfo{year}{2023}).
\newblock


\bibitem[Shi et~al\mbox{.}(2023)]%
        {shi2023replug}
\bibfield{author}{\bibinfo{person}{Weijia Shi}, \bibinfo{person}{Sewon Min}, \bibinfo{person}{Michihiro Yasunaga}, \bibinfo{person}{Minjoon Seo}, \bibinfo{person}{Rich James}, \bibinfo{person}{Mike Lewis}, \bibinfo{person}{Luke Zettlemoyer}, {and} \bibinfo{person}{Wen-tau Yih}.} \bibinfo{year}{2023}\natexlab{}.
\newblock \showarticletitle{Replug: Retrieval-augmented black-box language models}.
\newblock \bibinfo{journal}{\emph{arXiv preprint arXiv:2301.12652}} (\bibinfo{year}{2023}).
\newblock


\bibitem[Tian et~al\mbox{.}(2023)]%
        {tian2023askforcalibration}
\bibfield{author}{\bibinfo{person}{Katherine Tian}, \bibinfo{person}{Eric Mitchell}, \bibinfo{person}{Allan Zhou}, \bibinfo{person}{Archit Sharma}, \bibinfo{person}{Rafael Rafailov}, \bibinfo{person}{Huaxiu Yao}, \bibinfo{person}{Chelsea Finn}, {and} \bibinfo{person}{Christopher~D. Manning}.} \bibinfo{year}{2023}\natexlab{}.
\newblock \bibinfo{title}{Just Ask for Calibration: Strategies for Eliciting Calibrated Confidence Scores from Language Models Fine-Tuned with Human Feedback}.
\newblock
\newblock
\showeprint[arxiv]{2305.14975}~[cs.CL]


\bibitem[Wang et~al\mbox{.}(2023)]%
        {wang2023industrydomainqa}
\bibfield{author}{\bibinfo{person}{Zezhong Wang}, \bibinfo{person}{Fangkai Yang}, \bibinfo{person}{Pu Zhao}, \bibinfo{person}{Lu Wang}, \bibinfo{person}{Jue Zhang}, \bibinfo{person}{Mohit Garg}, \bibinfo{person}{Qingwei Lin}, {and} \bibinfo{person}{Dongmei Zhang}.} \bibinfo{year}{2023}\natexlab{}.
\newblock \bibinfo{title}{Empower Large Language Model to Perform Better on Industrial Domain-Specific Question Answering}.
\newblock
\newblock
\showeprint[arxiv]{2305.11541}~[cs.CL]


\bibitem[Xiong et~al\mbox{.}(2023)]%
        {xiong2023express}
\bibfield{author}{\bibinfo{person}{Miao Xiong}, \bibinfo{person}{Zhiyuan Hu}, \bibinfo{person}{Xinyang Lu}, \bibinfo{person}{Yifei Li}, \bibinfo{person}{Jie Fu}, \bibinfo{person}{Junxian He}, {and} \bibinfo{person}{Bryan Hooi}.} \bibinfo{year}{2023}\natexlab{}.
\newblock \showarticletitle{Can LLMs Express Their Uncertainty? An Empirical Evaluation of Confidence Elicitation in LLMs}.
\newblock \bibinfo{journal}{\emph{arXiv preprint arXiv:2306.13063}} (\bibinfo{year}{2023}).
\newblock


\bibitem[Ye et~al\mbox{.}(2023)]%
        {ye2023hallucinate}
\bibfield{author}{\bibinfo{person}{Hongbin Ye}, \bibinfo{person}{Tong Liu}, \bibinfo{person}{Aijia Zhang}, \bibinfo{person}{Wei Hua}, {and} \bibinfo{person}{Weiqiang Jia}.} \bibinfo{year}{2023}\natexlab{}.
\newblock \bibinfo{title}{Cognitive Mirage: A Review of Hallucinations in Large Language Models}.
\newblock
\newblock
\showeprint[arxiv]{2309.06794}~[cs.CL]


\bibitem[Zhang et~al\mbox{.}(2023b)]%
        {zhang2023snowball}
\bibfield{author}{\bibinfo{person}{Muru Zhang}, \bibinfo{person}{Ofir Press}, \bibinfo{person}{William Merrill}, \bibinfo{person}{Alisa Liu}, {and} \bibinfo{person}{Noah~A. Smith}.} \bibinfo{year}{2023}\natexlab{b}.
\newblock \bibinfo{title}{How Language Model Hallucinations Can Snowball}.
\newblock
\newblock
\showeprint[arxiv]{2305.13534}~[cs.CL]


\bibitem[Zhang et~al\mbox{.}(2023c)]%
        {zhang2023getting}
\bibfield{author}{\bibinfo{person}{Shizhuo~Dylan Zhang}, \bibinfo{person}{Talia Ringer}, {and} \bibinfo{person}{Emily First}.} \bibinfo{year}{2023}\natexlab{c}.
\newblock \bibinfo{title}{Getting More out of Large Language Models for Proofs}.
\newblock
\newblock
\showeprint[arxiv]{2305.04369}~[cs.FL]


\bibitem[Zhang et~al\mbox{.}(2023a)]%
        {zhang2023sirenshallucinatesurvey}
\bibfield{author}{\bibinfo{person}{Yue Zhang}, \bibinfo{person}{Yafu Li}, \bibinfo{person}{Leyang Cui}, \bibinfo{person}{Deng Cai}, \bibinfo{person}{Lemao Liu}, \bibinfo{person}{Tingchen Fu}, \bibinfo{person}{Xinting Huang}, \bibinfo{person}{Enbo Zhao}, \bibinfo{person}{Yu Zhang}, \bibinfo{person}{Yulong Chen}, \bibinfo{person}{Longyue Wang}, \bibinfo{person}{Anh~Tuan Luu}, \bibinfo{person}{Wei Bi}, \bibinfo{person}{Freda Shi}, {and} \bibinfo{person}{Shuming Shi}.} \bibinfo{year}{2023}\natexlab{a}.
\newblock \bibinfo{title}{Siren's Song in the AI Ocean: A Survey on Hallucination in Large Language Models}.
\newblock
\newblock
\showeprint[arxiv]{2309.01219}~[cs.CL]


\bibitem[Zhao et~al\mbox{.}(2023)]%
        {zhao2023paretocalibration}
\bibfield{author}{\bibinfo{person}{Theodore Zhao}, \bibinfo{person}{Mu Wei}, \bibinfo{person}{J~Samuel Preston}, {and} \bibinfo{person}{Hoifung Poon}.} \bibinfo{year}{2023}\natexlab{}.
\newblock \showarticletitle{Automatic Calibration and Error Correction for Large Language Models via Pareto Optimal Self-Supervision}.
\newblock \bibinfo{journal}{\emph{arXiv preprint arXiv:2306.16564}} (\bibinfo{year}{2023}).
\newblock


\bibitem[Zhou et~al\mbox{.}(2023)]%
        {zhou2023overconfidence}
\bibfield{author}{\bibinfo{person}{Kaitlyn Zhou}, \bibinfo{person}{Dan Jurafsky}, {and} \bibinfo{person}{Tatsunori Hashimoto}.} \bibinfo{year}{2023}\natexlab{}.
\newblock \bibinfo{title}{Navigating the Grey Area: Expressions of Overconfidence and Uncertainty in Language Models}.
\newblock
\newblock
\showeprint[arxiv]{2302.13439}~[cs.CL]


\bibitem[Ziegler et~al\mbox{.}(2019)]%
        {Ziegler2019humanpref}
\bibfield{author}{\bibinfo{person}{Daniel~M. Ziegler}, \bibinfo{person}{Nisan Stiennon}, \bibinfo{person}{Jeff Wu}, \bibinfo{person}{Tom~B. Brown}, \bibinfo{person}{Alec Radford}, \bibinfo{person}{Dario Amodei}, \bibinfo{person}{Paul Christiano}, {and} \bibinfo{person}{Geoffrey Irving}.} \bibinfo{year}{2019}\natexlab{}.
\newblock \showarticletitle{Fine-Tuning Language Models from Human Preferences}.
\newblock \bibinfo{journal}{\emph{ArXiv}}  \bibinfo{volume}{abs/1909.08593} (\bibinfo{year}{2019}).
\newblock
\urldef\tempurl%
\url{https://api.semanticscholar.org/CorpusID:202660943}
\showURL{%
\tempurl}


\end{thebibliography}

\appendix









\end{document}